\documentclass[10pt,twocolumn,letterpaper]{article}

\usepackage[pagenumbers]{iccv} 

\usepackage[dvipsnames]{xcolor}
\usepackage{amsmath}
\usepackage{amsthm}
\usepackage{amsfonts}
\usepackage{amssymb}
\usepackage{graphicx}
\usepackage{adjustbox}
\usepackage{multicol}
\usepackage{mathtools}
\usepackage{bm}
\usepackage{caption}
\usepackage{subcaption}
\usepackage{multirow}
\usepackage{tikz}
\usepackage[shortlabels]{enumitem}
\definecolor{cvprblue}{rgb}{0.21,0.49,0.74}
\usepackage[pagebackref,breaklinks,colorlinks,allcolors=cvprblue]{hyperref}

\def\paperID{11517} 
\def\confName{ICCV}
\def\confYear{2025}
\def\heroFigure{\vspace{0.5em}
\begin{minipage}{\textwidth}
\centering
\captionsetup{type=figure}
\includegraphics[width=0.7\textwidth]{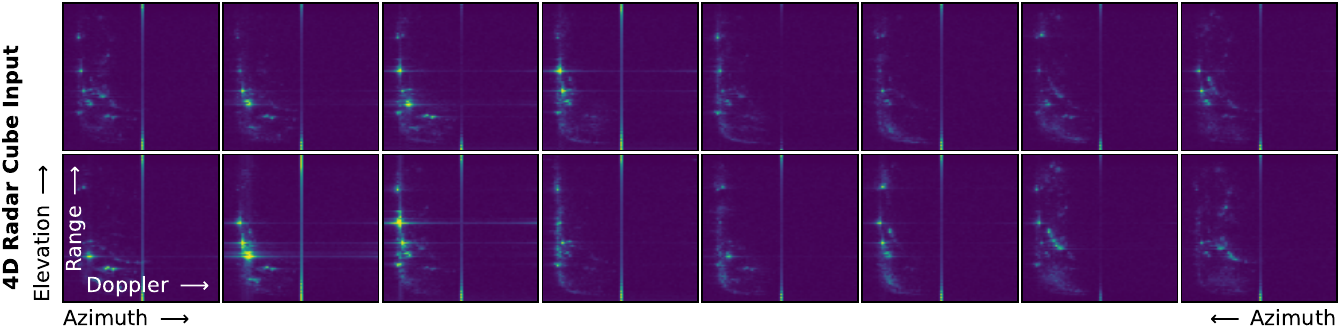}
\includegraphics[width=1.0\textwidth]{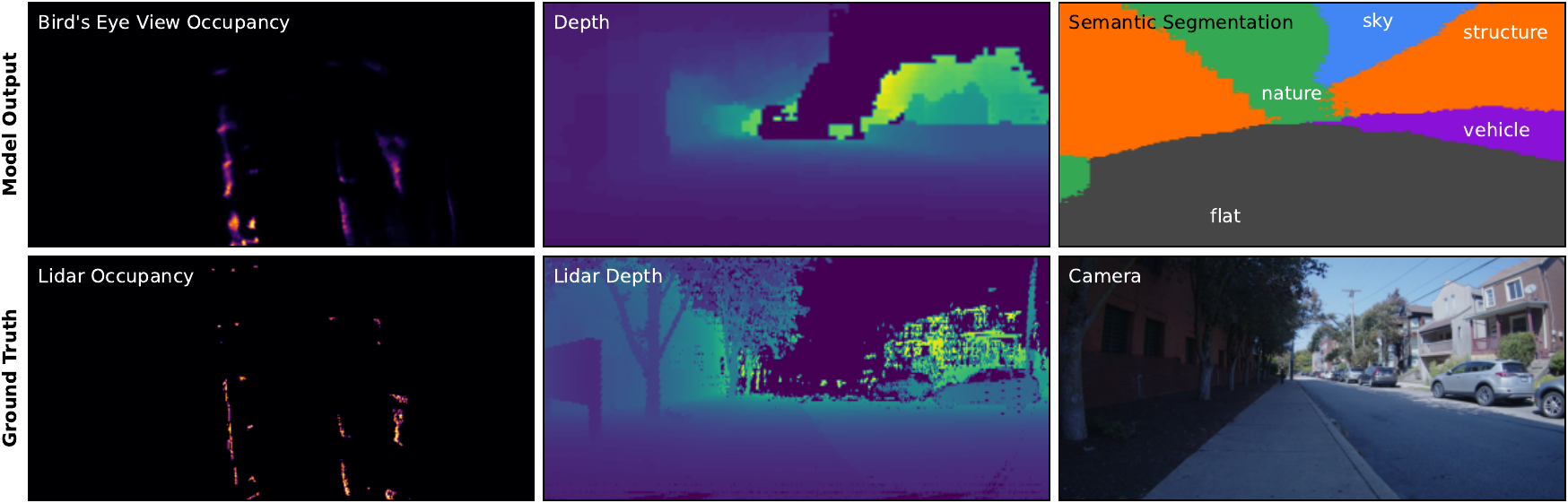}
\vspace{-1.8em}
\captionof{figure}{\textbf{Using only 4D range-Doppler-azimuth-elevation data from a radar with a 3x4 antenna array -- equivalent to a 0.26-megapixel camera,} we trained (separate) radar transformers on 24 hours (1M radar-lidar-camera samples) of data to predict Bird's Eye View 2D occupancy (left), 3D occupancy (center), semantic segmetation (right), and Ego-Motion.}
\label{fig:hero}
\end{minipage}
\vspace{-1em}
}

\title{Towards Foundational Models for Single-Chip Radar}

\author{
Tianshu Huang\textsuperscript{12}
\hspace{1.5em} Akarsh Prabhakara\textsuperscript{3}
\hspace{1.5em} Chuhan Chen\textsuperscript{1}
\hspace{1.5em} Jay Karhade\textsuperscript{1}
\\
\hspace{1.5em} Deva Ramanan\textsuperscript{1}
\hspace{1.5em} Matthew O'Toole\textsuperscript{1}
\hspace{1.5em} Anthony Rowe\textsuperscript{12}
\\
\textsuperscript{1}Carnegie Mellon University
\hspace{1.5em}\textsuperscript{2}Bosch Research
\hspace{1.5em}\textsuperscript{3}University of Wisconsin--Madison
\\
{\tt\small \{tianshu2, chuhanc, jkarhade, deva, motoole2, agr\}@andrew.cmu.edu, akarsh@cs.wisc.edu}
}

\begin{document}
\maketitle

\begin{abstract}
mmWave radars are compact, inexpensive, and durable sensors that are robust to occlusions and work regardless of environmental conditions, such as weather and darkness. However, this comes at the cost of poor angular resolution, especially for inexpensive single-chip radars, which are typically used in automotive and indoor sensing applications. Although many have proposed learning-based methods to mitigate this weakness, no standardized foundational models or large datasets for the mmWave radar have emerged, and practitioners have largely trained task-specific models from scratch using relatively small datasets.

In this paper, we collect (to our knowledge) the largest available raw radar dataset with 1M samples (29 hours) and train a foundational model for 4D single-chip radar, which can predict 3D occupancy and semantic segmentation with quality that is typically only possible with much higher resolution sensors. We demonstrate that our Generalizable Radar Transformer (GRT) generalizes across diverse settings, can be fine-tuned for different tasks, and shows logarithmic data scaling of 20\% per $10\times$ data. We also run extensive ablations on common design decisions, and find that using raw radar data significantly outperforms widely-used lossy representations, equivalent to a $10\times$ increase in training data. Finally, we roughly estimate that $\approx$100M samples (3000 hours) of data are required to fully exploit the potential of GRT.
\end{abstract}
\vspace{-0.5em}
\section{Introduction}

As a compact, inexpensive \cite{lien2016soli}, and robust solid-state sensor, mmWave radars are ideal for sensing applications ranging from simple automatic door openers \cite{ti-dooropener} to autonomous drones \cite{doer2021yaw} or vehicles \cite{russell1997millimeter,waldschmidt2021automotive}. mmWave radars are
rich sensors which can directly measure range and velocity while capturing a unique range of material properties \cite{huang2024dart}; however, this comes at the cost of poor angular resolution typically on the order of $15^\circ$ -- orders of magnitude worse than cameras or lidars \cite{ti:AWR1843AOP}.

\begin{figure}
\centering
\includegraphics[width=\columnwidth]{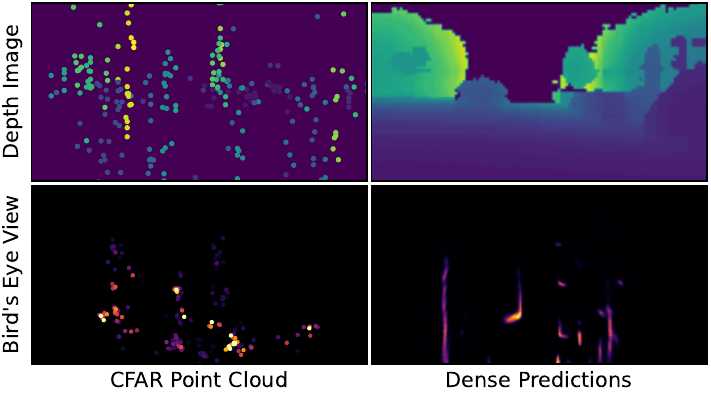}
\vspace{-2em}
\caption{While a transformer can generate Lidar-like depth and bird's eye view images, traditional CFAR point clouds are noisy, and have poor angular resolution -- especially in the elevation axis.}
\label{fig:cfar_example}
\end{figure}

Radar data are typically processed into radar point clouds (Fig.~\ref{fig:cfar_example}) derived using Constant False Alarm Rate (CFAR) peak detectors \cite{minkler1990cfar,rohling1983radar} combined with Angle-of-Arrival estimation techniques \cite{vasanelli2020calibration}. However, this is a substantially lossy process: while raw radar data suffers from unique noise patterns such as ``bleed'' and side lobes \cite{li2021signal}, weak reflectors and other signals can be hidden in this noise, which would ordinarily be filtered out.

On the other hand, raw spectrum (4D range-Doppler-azimuth-elevation data cubes \cite{richards2005fundamentals}) can be unintuitive and difficult to interpret compared to lidar point clouds or camera images, and include properties such as specularity and Doppler which lack straight-forward Cartesian interpretations \cite{huang2024dart}. As such, many machine learning methods have been proposed \cite{zhang2021raddet,lai2024enabling,prabhakara2023high} to exploit 4D radar data from single-chip radars, achieving remarkable performance on 2D scene understanding tasks. However, due to the dominance of CFAR point clouds in radar processing, as well as the high data rate of raw mmWave radar data, most radar toolchains only process point-cloud data. Tooling for raw I/Q (in-phase/quadrature) data is often brittle, poorly documented, and largely unsupported by radar vendors, severely limiting the availability and scale of both raw mmWave datasets and the models which operate on raw data.

To rectify this limitation, we develop an open-source toolchain and associated large dataset specifically for 4D mmWave radar data. Training a radar-to-lidar model and fine-tuning for a range of other tasks, we demonstrate the surprising
effectiveness of mmWave radar models trained at scale (Fig.~\ref{fig:hero}). Going further, just as large foundational models \cite{bommasani2021opportunities} have greatly accelerated the pace of innovation in computer vision and natural language processing, we believe that a foundational model for raw mmWave radar trained at even larger scale could similarly supercharge the advancement of radar sensing techniques.

\vspace{-1.5em}
\paragraph{Contributions}
In this paper, we develop a full stack\footnote{
Our data collection system, dataset, code, and model can be found via our project site: \url{https://wiselabcmu.github.io/grt/}.
} for collecting data, training, and evaluating a transformer for 4D single-chip radar to quantify both the potential costs and benefits of training a foundational model at scale. To summarize our contributions:
\begin{enumerate}[(1)]
    \item We develop a compact, lightweight multimodal data collection system (Sec.~\ref{sec:data_collection}) capable of collecting synchronized raw radar, Lidar, and camera data which can be operated as a handheld device. Our system can be easily replicated using off-the-shelf components, 3D printed parts, and our open-source software.
    \item Using this data collection system, we collect a dataset, I/Q-1M (\textit{One Million IQ Frames}), consisting of 29 hours of data -- 8$\times$ longer than the next largest publicly available raw radar dataset -- split between indoor, outdoor-handheld, and bike-mounted settings, each with different radar configurations (Sec.~\ref{sec:collected_data}).
    \item Finally, using our dataset, we train a Generalizable Radar Transformer (GRT), which can output depth maps and segmentation images with quality which is typically only possible with much higher resolution radars. Using GRT, we then run ablations on common design choices (Sec.~\ref{sec:ablations}), quantify the scalability of GRTs with increasing dataset and model size (Sec.~\ref{sec:foundational_model}, \ref{sec:how_much_data}), and demonstrate that our GRT can be readily fine-tuned for other tasks and settings (Sec.~\ref{sec:tuning}), including obtaining state-of-the-art performance on the Coloradar \cite{kramer2022coloradar} dataset with 30-minutes of fine-tuning.
\end{enumerate}

\vspace{-0.8em}
\paragraph{Key Findings}

We summarize our key findings as follows:
\begin{itemize}
    \item \textbf{Radar models can generalize} to different settings and radar configurations (Sec.~\ref{sec:foundational_model}), as well as across objectives (with some fine-tuning). This suggests great potential for a cross-domain foundational model to improve and accelerate the development of new radar models.
    \item \textbf{Using raw data yields outsized performance gains}, equivalent to more than a $10\times$ increase in training data (Sec.~\ref{sec:ablations}). While existing datasets largely focus on CFAR point clouds or other processed representations, we believe that more emphasis should be placed on making raw data available for research.
    \item \textbf{Existing mmWave radar datasets are vastly undersized}. 24 hours of training data is not enough to saturate even a 4M parameter model! Our analysis suggests that \textbf{at least 100$\times$ more data} is required to exploit the full potential of radar transformer models (Sec.~\ref{sec:how_much_data}).
\end{itemize}

\vspace{-1em}
\paragraph{Limitations} Despite its size compared to previous datasets, I/Q-1M is quite small compared to the datasets used to train modern vision transformers, which can exceed billions of samples \cite{zhai2022scaling}.
I/Q-1M also only includes daylight conditions and fair weather, and lacks the scale to capture ``edge cases'' that would be represented in a larger dataset \cite{kalra2016driving}. Finally, since I/Q-1M uses a single type of radar, we cannot evaluate generalization across different antenna configurations --- only radar configurations.

\section{Related Work}


\begin{figure}
\centering
\includegraphics[width=\columnwidth]{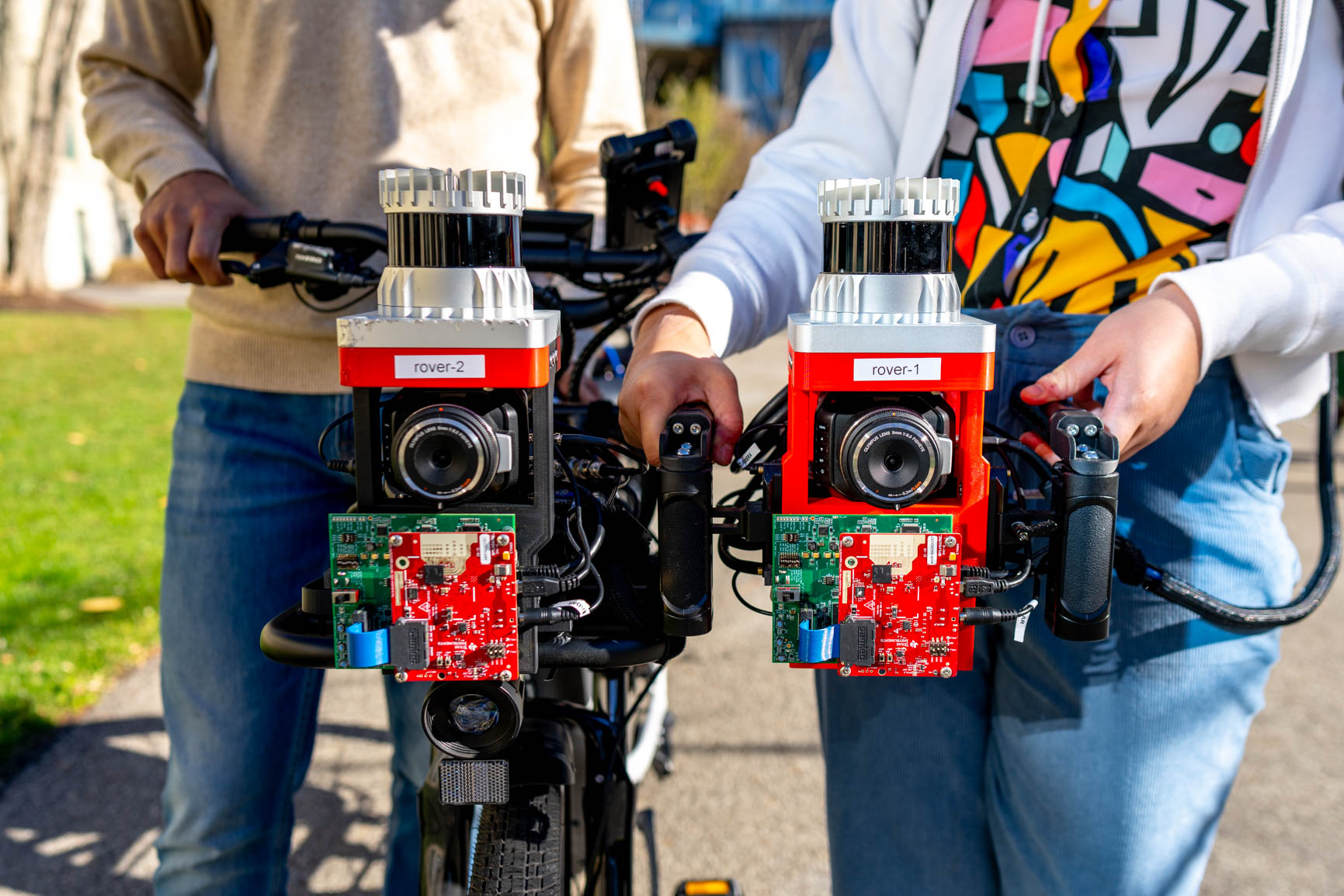}
\vspace{-1.7em}
\caption{\textbf{Our data collection rig} in its handheld (right) and bike-mounted (left) configurations; see App.~\ref{app:dataset} for additional images.}
\vspace{-0.5em}
\label{fig:rover}
\end{figure}

\paragraph{4D Solid State Radar} Excluding mechanical radars, which perceive the world as lidar-like 2D heatmaps \cite{barnes2020oxford,sheeny2021radiate,burnett2023boreas} via a rotating antenna, solid-state mmWave radars operate by transmitting and receiving a sequence of frequency-modulated ``chirps'' from an array of transmit (TX) and receive (RX) antenna \cite{iovescu2017fundamentals}; this data is typically (losslessly) transformed to a 4D range-Doppler-azimuth-elevation data cube using a 4D FFT \cite{richards2005fundamentals}, whose resolutions are constrained by bandwidth, form factor, and the integration window. We focus on single-chip radars which have compact form factors -- and thus poor angular resolution.

\vspace{-1em}
\paragraph{Learning and Datasets for 4D Radar} Most radar processing methods use point clouds extracted from the 4D cubes \cite{minkler1990cfar,rohling1983radar} as inputs \cite{prabhakara2022exploring,armanious2024bosch,cui2024milipoint,lu2020milliego,shuai2021millieye,singh2019radhar}, allowing them to re-use popular Lidar architectures or even pre-trained models such as PointNet \cite{qi2017pointnet}. However, since this discards much of the information contained in a 4D radar cube, many competing approaches propose to directly interpret the 4D radar cube using methods and architectures such as feedforward convolutional architectures \cite{decourt2022darod,rebut2022raw,nowruzi2020deep,zhang2021raddet,palffy2020cnn}, multi-view convolutional networks across different tensor axes \cite{ouaknine2021multi,gao2020ramp,major2019vehicle}, U-Nets \cite{prabhakara2023high,li2023azimuth,orr2021high}, diffusion models \cite{zhang2024towards}, and transformers \cite{giroux2023t,bai2021radar,jin2023semantic,jin2023semantic}.

These prior machine learning-based approaches rely on publicly available datasets with 4D radar data, including from both cascaded \cite{paek2022k,rebut2022raw} and single-chip \cite{lim2021radical,kramer2022coloradar,zhang2021raddet} radars; however, existing datasets are relatively small, with 3.6-hour RaDICal \cite{lim2021radical} and 2.4-hour Coloradar \cite{kramer2022coloradar} as the largest (Table~\ref{tab:other_datasets_short}). Due to the success of powerful but data-hungry \cite{zhai2022scaling} transformer models \cite{vaswani2017attention} in computer vision \cite{dosovitskiy2021an}, we believe that limited data availability imposes a substantial bottleneck on learning for 4D radar.

\vspace{-1em}
\paragraph{High-resolution Imaging from Low-Resolution Radar} Due to the low angular resolution of single-chip radars, extracting high-resolution angular information can be challenging; as a result, prior work focuses on recovering 2D spatial information \cite{prabhakara2023high,giroux2023t,zhang2024towards}. Thus, while prior methods can extract 3D information using high-resolution cascaded radars \cite{ding2025radarocc}, 3D imaging from single-chip radars generally requires additional information such as structured motion or multiple views, for example segmentation and maps using a rotating single-chip radar \cite{lai2024enabling}, a 3D occupancy map using multiple views \cite{huang2024dart}, or high resolution images from fixed trajectories using Synthetic Aperture Radar \cite{gao2021mimo,mamandipoor201460,yamada2017high,yanik2019near}. Instead, we show that by leveraging a sufficiently large dataset, even single frames are sufficient to recover dense angular resolution in both azimuth and elevation.


\section{Data Collection System and Dataset}

\begin{table}
\caption{\textbf{Comparison with other single-chip mmWave radar datasets;} a \textit{frame} refers to the number of unique radar-sensor samples. For comparisons with datasets using other types of radar, see App.~\ref{app:other_datasets}. Our dataset is significantly larger than previous single-chip radar datasets, enabling us to explore scaling up models.}
\vspace{-0.7em}
\centering
\footnotesize
\begin{tabular}{ccccc}
\toprule
Dataset & 4D Data Cube & Dataset Size \\
\toprule
\textbf{I/Q-1M (Ours)} & \textbf{Yes} & \textbf{29 hours (1M frames)} \\
MilliPoint \cite{cui2024milipoint} & No (3D Points) & 6.3 Hours (545k frames) \\
RaDICal \cite{lim2021radical} & \textbf{Yes} & 3.6 Hours (394k frames) \\
CRUW \cite{wang2021rodnet} & No (2D Map) & 3 hours (400k frames) \\
Coloradar \cite{kramer2022coloradar} & \textbf{Yes} & 2.4 hours (82k frames) \\
RadarHD \cite{prabhakara2023high} & \textbf{Yes} & 200k frames \\
CARRADA \cite{ouaknine2021carrada} & No (3D Cube) & 21 Minutes (13k frames) \\
RADDet \cite{zhang2021raddet} & \textbf{Yes} & 10k frames \\
\toprule
\end{tabular}
\label{tab:other_datasets_short}
\end{table}

Dataset scale is key to training and evaluating potential foundational models. As such, we developed a scalable data collection system (Fig.~\ref{fig:rover}), which we used to collect a large raw mmWave radar dataset, consisting of 1M radar-lidar-camera samples over 29 hours (Table~\ref{tab:other_datasets_short}). For additional details on our dataset and data collection rig, see App.~\ref{app:dataset}.

\subsection{Data Collection System}
\label{sec:data_collection}

Our data collection system, \texttt{red-rover}, was built around a TI AWR1843 Radar, Lidar, Camera, and IMU which can be easily operated via a simple web app on a mobile phone. Our system records all data to a single hot-swappable external drive via a single linux computer which handles time synchronization, minimizing turnaround time.

We also designed our data collection system to have a compact, battery-operated form factor to allow for a variety of collection modalities, including handheld and bicycle mounted. This also allows us to collect data relevant for tasks such as indoor sensing and localization, which are underrepresented in existing datasets, while still collecting automotive-like data by mounting our system to an E-bike.

\subsection{Collected Data}
\label{sec:collected_data}

\begin{table}
\caption{\textbf{Key specifications for each setting.} Settings have varying max Doppler $D_{\text{max}}$ and range $R_{\text{max}}$; all traces used a fixed resolution of 64 Doppler and 256 range bins.}
\vspace{-0.7em}
\centering
\footnotesize
\begin{tabular}{ccccccc}
\toprule
Setting & Size & Length & Average Speed & $D_{\text{max}}$ &  $R_{\text{max}}$ \\
\toprule
\texttt{indoor} & 310k & 8.9h & 1.0m/s & 1.2m/s & 11.2m \\
\texttt{outdoor} & 372k & 10.7h & 1.4m/s & 1.8m/s & 22.4m \\
\texttt{bike} & 333k & 9.3h & 5.4m/s & 8.0m/s & 22.4m\\
\toprule
\end{tabular}
\vspace{-1em}
\label{tab:splits}
\end{table}

\begin{figure*}
\includegraphics[width=\textwidth]{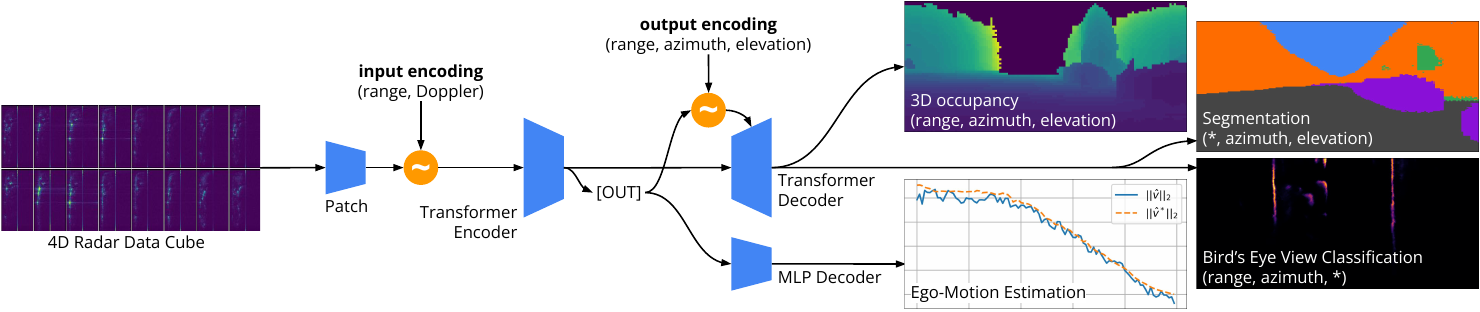}
\vspace{-2em}
\caption{\textbf{The GRT architecture.} 4D radar cubes are patched with a linear projection, and a sinusoidal positional encoding is added. A transformer architecture is then used, with a transformer decoder for dense outputs and a MLP decoder for Ego-Motion estimation; different output encodings are used depending on the output axes and resolution.}
\vspace{-1em}
\label{fig:method}
\end{figure*}

We collected three roughly equally sized splits (Table~\ref{tab:splits}) from indoor handheld, outdoor handheld, and bike-mounted settings on the CMU campus and Pittsburgh area:
\begin{itemize}
    \item \texttt{indoor}: inside buildings at a slow to moderate walking pace, visiting multiple floors and areas within each.
    \item \texttt{outdoor}: neighborhoods ranging from single family detached to high density commercial zoning at a moderate to fast walking pace.
    \item \texttt{bike}: bike rides in different directions from a set starting point with a moderate biking pace.
\end{itemize}
Each setting features a mobile observer, with radar modulation parameters tuned for typical speeds. For sample data from each setting, see App.~\ref{app:settings}.

\section{Methodology}

\begin{table}
\caption{\textbf{Transformer sizes.} \emph{Layers} indicates the number of \texttt{encoder + decoder} layers; \emph{Speed} indicates the (batched) inference throughput of each model on a single RTX 4090.}
\vspace{-0.7em}
\centering
\footnotesize
\begin{tabular}{ccccc}
\toprule
Size & Layers & Dimension & Params & Speed \\
\toprule
\texttt{pico} & 2 + 2 & 256 (4 heads) & 3.9M & 750 fps \\
\texttt{tiny} & 3 + 3 & 384 (6 heads) & 12.7M & 320 fps \\
\texttt{small} & 4 + 4 & 512 (8 heads) & 28.9M & 170 fps \\
\texttt{medium} & 6 + 6 & 640 (10 heads) & 69.4M & 84 fps \\
\texttt{large} & 9 + 9 & 768 (12 heads) & 149M & 44 fps \\
\toprule
\end{tabular}
\label{tab:sizes}
\end{table}

Using a transformer architecture (Sec.~\ref{sec:architecture}), we train our Generalizable Radar Transformer (GRT) for a range of different tasks (Sec.~\ref{sec:base_task}-\ref{sec:other_tasks}), and evaluate it on our dataset using a rigorous statistical methodology (Sec.~\ref{sec:eval_methodology}).

\subsection{Model Architecture}
\label{sec:architecture}

While many architectural refinements exist for vision transformers \cite{liu2021swin,ranftl2021vision}, as well as for radar specifically \cite{li2023azimuth,giroux2023t}, we use a direct adaptation (Fig.~\ref{fig:method}) of the original Transformer \cite{vaswani2017attention} and Vision Transformer \cite{dosovitskiy2021an} to focus on measuring the fundamental properties of Radar transformers.

\paragraph{Radar Processing} From the (slow time, TX, RX, fast time) I/Q stream, we perform a 4-Dimensional FFT to obtain (range, Doppler, azimuth, elevation) dense 4D radar data cubes of size (256, 64, 8, 2), which we provide to the model as two channels consisting of the amplitude and phase angle. This data cube is patched along the range and Doppler dimensions into patches of size $4 \times 2(\times 8 \times 2)$, yielding an initial set of $64 \times 32 = 2048$ patch tokens.

\vspace{-0.9em}
\paragraph{Transformer Architecture} Crucially, unlike a vision transformer \cite{dosovitskiy2021an}, radar models take inputs that have different \textit{spatial axes} than their outputs, with vastly different relative resolutions where they overlap. As such, we use a standard transformer \textit{with a decoder} \cite{vaswani2017attention}, with varying layers and widths (Table~\ref{tab:sizes}); for a full specification of our transformer architecture and training procedure, see App.~\ref{app:hyperparameters}.

\vspace{-0.9em}
\paragraph{Decoder Query} To handle the ``change of basis'' between the input and output space, we use an architecture based on Perceiver I/O \cite{jaegle2021perceiver}. We start by concatenating a (learned) output token to the encoder (similar to standard vision transformer \cite{darcet2023vision}). The encoder output corresponding to the output token is then tiled into the desired decoder shape with a 3D sinusoidal positional encoding applied and is used as the input to the decoder, which attends to the encoder outputs.

\subsection{Base Task: 3D Occupancy Classification}
\label{sec:base_task}

An ideal foundational model training task should be easy to gather data for (e.g., using self-supervised learning) and closely aligned with a wide range of potential downstream tasks. As such, since we are primarily concerned with understanding the \textit{spatial} relationship between 4D radar data and 3D space, we use 3D (polar) occupancy classification -- predicting the occupancy
of $64 \times 128 \times 64$ range-azimuth-elevation cells -- as a base task, with Lidar as a ground truth. Our task uses a binary cross entropy objective, with some weighting to correct for cell sizes; see App.~\ref{app:tasks} for details.

Notably, in addition to being fully self-supervised, this task covers all three possible \textit{output dimensions} (range, azimuth, and elevation), meaning that downstream tasks such as range-azimuth classification or azimuth-elevation segmentation can be cast as 2D slices of this 3D output. This enables us to fine-tune for tasks, even if they have different spatial dimensions, simply by replacing the output head and modifying the output positional encoding queries.

\subsection{Other Tasks}
\label{sec:other_tasks}

In order to evaluate GRT's suitability as a \textit{foundational model} for downstream fine-tuning, we use three additional tasks, each representing different output dimensions:
\begin{itemize}
    \item \textbf{Bird's Eye View (BEV) Occupancy}: Similarly to \cite{prabhakara2023high,giroux2023t}, we classify the $256 \times 1024$ \textbf{range-azimuth} polar occupancy using Lidar as a ground truth, with the range normalized to the radar's range resolution.
    \item \textbf{Semantic Segmentation}: Similar to \cite{lai2024enabling}, we train our GRT to output $640\times 640$ \textbf{azimuth-elevation} class labels. Since radars cannot feasibly identify many classes (e.g., poster vs. sign vs. wall) which a camera could, we use eight coarse categories: person, sky, vehicle, flat, nature, structure, ceiling, and object.
    \item \textbf{Ego-Motion Estimation}: Since radars can ``natively'' measure velocity\footnote{Sensors which provide ``absolute'' pose, e.g. Lidars and Cameras, must differentiate, while IMUs must integrate.}, we predict the velocity of the radar relative to its current orientation. Since ego-motion estimation does not require a dense output, we replace the transformer decoder with a multi-layer-perceptron decoder head with 3 layers of 512 units.
\end{itemize}
For more detail about each task, see App.~\ref{app:tasks}.

\subsection{Evaluation}
\label{sec:eval_methodology}

Due to the cost of scaling foundational models, false positives can result in significant wasted resources, especially if associated with a costly methodological change. As such, since our dataset cannot be treated as having an ``infinite sample size'', we calculate upper-bounded uncertainty estimates wherever possible. In order to ensure these results are statistically accurate, we take the following steps:
\begin{itemize}
    \item \textbf{Geo-Split}: Within each setting, $\approx$1.5 hours of data was reserved as a test set, which we ensured to be geographically disjoint from the training set to prevent leakage \cite{lilja2024localization}.
    \item \textbf{Sample Size correction}: Time series signals -- e.g. radar-lidar-video tuples -- cannot be viewed as independent samples; as such, the \textit{effective sample size}, which we obtain from an autocorrelation-based estimate \cite{robert1999monte}, must be used when calculating the standard error.
    \item \textbf{Paired z-Test}: Using the fact that models are evaluated on the same test traces, we use a paired z-test on the relative performance of each model with respect to a baseline.
\end{itemize}
We report each metric relative to its specified baseline by default, along with error bars for a two-sided 95\% confidence interval; using our procedure and dataset, we can measure differences of 1-2\% (App.~\ref{app:procedure}).

\vspace{-0.8em}
\paragraph{Validation Split and Data Size Sampling.} We used the last 10\% of each training trace for validation (separate from the test set), with the first 90\% used for training. When training on reduced dataset sizes, we use the first 9\%, 18\%, and 45\% of each trace for training for 10\%, 20\%, 50\% dataset sizes respectively; to reduce the variance of our experiments, we always 10\% of each trace for validation.

\section{Results}
\label{sec:results}

\begin{figure*}
\centering
\includegraphics[width=\textwidth]{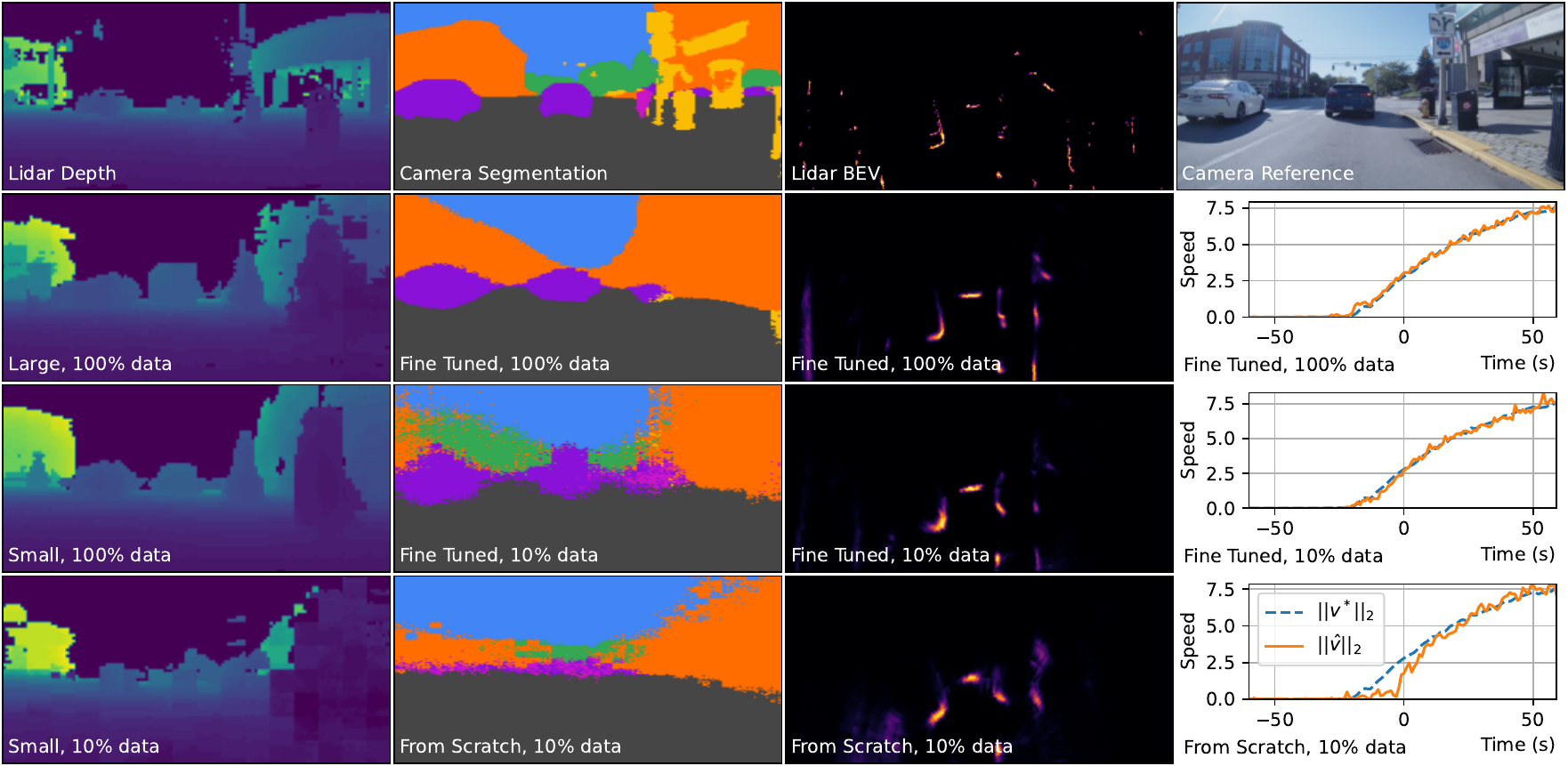}
\vspace{-2em}
\caption{\textbf{Training with more data and fine tuning instead of starting from scratch lead to much higher performance.} 3D occupancy maps are less noisy as seen in rendered depth images (left), semantic segmentation is cleaner (center left), bird's eye view occupancy is sharper (center right), and velocity estimation is more accurate (right). However, scaling model size (left) does not have a large impact.}
\vspace{-0.5em}
\label{fig:scaling_examples}
\end{figure*}

Using our dataset, we first ran extensive ablation (Sec.~\ref{sec:ablations}), scaling (Sec.~\ref{sec:foundational_model}), and fine-tuning (Sec.~\ref{sec:tuning}) experiments which show the efficacy, scalability, and generalizability of GRT. Our experiments took 874 RTX 4090-hours\footnote{
Our experiments were run on a range of different machines with varying compute capacity, which we normalize with respect to a single RTX 4090. We only tracked training and validation time, with testing excluded.
} of training time, with the \texttt{GRT-small} model taking 22 RTX 4090-hours to train.

\vspace{-1em}
\paragraph{Model Performance}

Despite the low resolution (only 3 TX $\times$ 4 TX antenna) of our radar, GRT is able to predict a range of outputs with remarkable quality (Fig.~\ref{fig:scaling_examples}); we show additional examples in App.~\ref{app:result_samples}. We also evaluated common metrics for each objective as an absolute reference (App.~\ref{app:metrics}); GRT achieves a 3D chamfer distance of 4.9 range bins, corresponding to 0.66m \texttt{indoors}, 1.6m \texttt{outdoors}, and 1.5m on \texttt{bike}.

\subsection{Ablation Studies}
\label{sec:ablations}

Using our dataset and GRT, we performed ablation studies on several parameters that are independent of the underlying architecture and the task (Table~\ref{tab:ablations}). In particular, we find that several common practices -- omitting Doppler information, using Angle-of-Arrival Estimates, and applying CFAR thresholding -- result in degraded performance equivalent to more than a $10\times$ reduction in training data.

\vspace{-1em}
\paragraph{Input Representation} Our models use complex 4D radar cubes that losslessly capture all information measured by a radar; however, the common practice in radar models and data sets is to use processed higher-level representations. We benchmark three common approaches:
\begin{itemize}
    \item \textbf{Real (amplitude-only) 4D data cubes} are not significantly different from complex data, indicating that the ``leftover'' phase from a Doppler FFT carries little additional information. Since using complex data has negligible compute overhead, we default to complex inputs.
    \item \textbf{Angle of Arrival Estimates} can be used to replace dense antenna measurements. Since our radar has only 2 elevation bins, we reduce the azimuth axis (8 bins) into an AoA estimate.
    This discards a substantial amount of information, leading to a 28.0\% loss increase, equivalent to more than a 10$\times$ decrease in the size of the dataset.
    \item \textbf{Constant False Alarm Rate (CFAR)} processing removes weak or noisy reflectors based on local estimates of background noise \cite{minkler1990cfar}; this ablation uses a p-value threshold of 0.05, and zeros out rejected points. This leads to an even higher 31.5\% loss increase.
\end{itemize}

\begin{table}
\footnotesize
\caption{\textbf{Test loss for each ablation} (smaller is better) relative to \texttt{GRT-small} trained on our full dataset, along with 95\% confidence intervals for the relative differences.}
\vspace{-0.7em}
\begin{tabular}{llc}
\toprule
\multicolumn{2}{c}{Ablation} & Relative Test Loss \\
\toprule
Inputs & Amplitude Only & $+0.04 \pm 0.85\%$ \\
    & Angle of Arrival & $+28.0 \pm 2.12\%$ \\
    & CFAR Thresholding & $+31.5 \pm 2.38\%$ \\
\hline
Doppler & Without Doppler FFT & $+22.5 \pm 1.76\%$ \\
    & Slow Time Shuffled & $+23.10 \pm 1.94\%$ \\
\hline
Post-Patch Axes & Doppler-Az-El & $+6.22 \pm 1.11\%$ \\
    & Range-Az-El & $+6.27 \pm 1.10\%$ \\
    & Range-Doppler-Az-El & $+4.18 \pm 1.09\%$ \\
\hline
Augmentations & None & $+5.87 \pm 1.32\%$ \\
    & Scale, Phase, and Flip Only & $+3.89 \pm 1.19\%$ \\
\hline
Separate Models & Indoor Data Only & $+5.76 \pm 1.85\%$ \\
    & Outdoor Data Only & $+5.58 \pm 1.26\%$ \\
    & Bike Data Only & $+2.77 \pm 1.41\%$ \\
\toprule
\end{tabular}
\label{tab:ablations}
\end{table}

\begin{figure*}
\centering
\includegraphics[width=\textwidth]{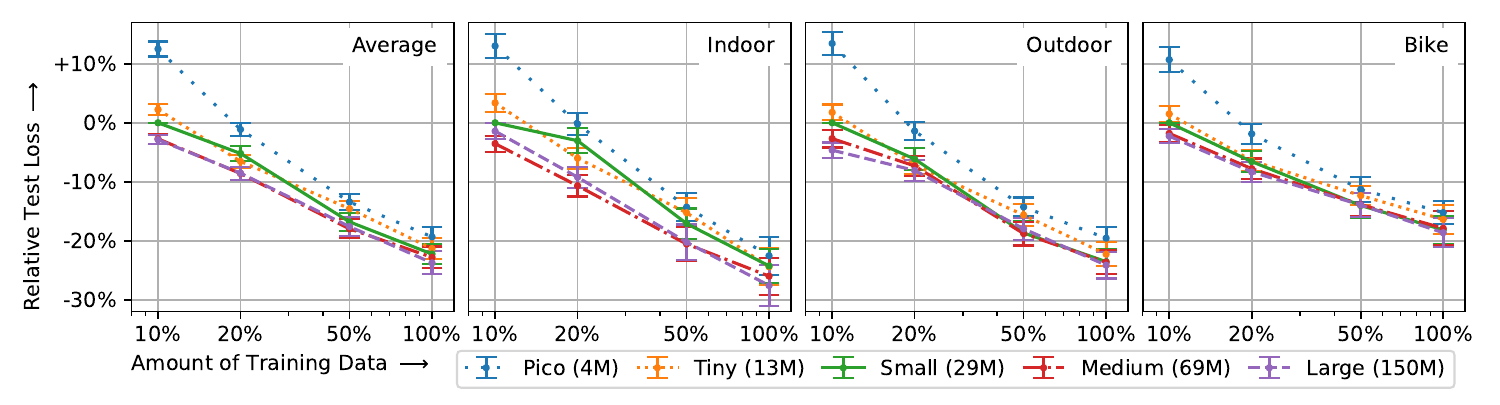}
\vspace{-2.5em}
\caption{\textbf{Scaling laws for mmWave radar transformers} across \texttt{indoor}, \texttt{outdoor}, and \texttt{bike} test splits. \textmd{Models and confidence intervals are measured relative to the \texttt{small} transformer trained on 10\% of the dataset. While our models show weak scaling over model size when trained on our dataset, we see strong log-linear scaling across dataset size of $\approx\!20\%$ loss decrease per 10$\times$ increase in data size. For variants of this graph with respect to absolute metrics, see App.~\ref{app:metrics}.}}
\label{fig:scaling}
\end{figure*}

\vspace{-1em}
\paragraph{Impact of Doppler} With limited angular resolution, GRT is highly dependent on Doppler information, which can capture higher resolution geometry \cite{huang2024dart}. We find that removing the Doppler FFT from our processing pipeline (i.e. treating each 4D radar cube as a time series of 64 3D frames \cite{prabhakara2023high}) leads to a 22.5\% loss increase. As an additional ablation, we also shuffle the slow-time axis to fully destroy any Doppler information; this does not lead to a futher significant loss increase, suggesting that off-the-shelf transformers cannot easily learn FFTs. Finally, we observe worse performance at low speeds since less Doppler information is available at slow speeds (App.~\ref{app:doppler}).

\vspace{-1em}
\paragraph{Patch Axes} Since 4D range-Doppler-azimuth-elevation radar data cubes have four axes with different properties, they do not have an obvious counterpart to the square patches used in Vision Transformers. Benchmarking four alternatives (App.~\ref{app:hyperparameters}), with each resulting in 2048 total patches, we find that Range-Doppler patching where the azimuth and elevation axes are ``patched out'' (similar to \cite{giroux2023t,rebut2022raw}) is the most effective, performing $\approx$5\% better.

\vspace{-1em}
\paragraph{Data Augmentations}
We develop a range of data augmentations that together provide a modest but significant performance improvement ($5.87 \pm 1.32\%$; Table~\ref{tab:ablations}); we provide details about each augmentation in App.~\ref{app:augmentations}.

\subsection{Towards a Radar Foundational Model}
\label{sec:foundational_model}

\paragraph{Scalability vs. Baselines} 

\begin{figure}
    \vspace{-1em}
    \includegraphics[width=\columnwidth]{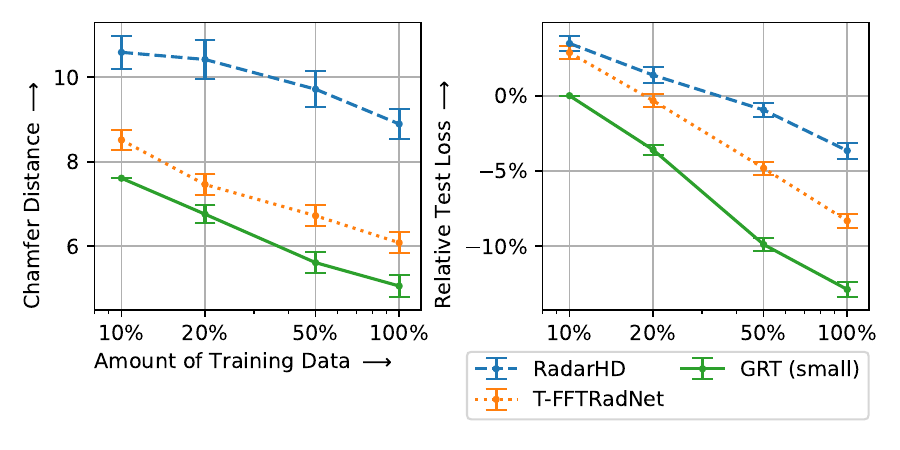}
    \vspace{-2.5em}
    \caption{\textbf{GRT compared to Baselines}, measured with respect to Chamfer distance (in range bins) and test loss across different training set sizes, averaged across our dataset.}
    \vspace{-0.5em}
    \label{fig:baselines}
\end{figure}

Since prior work on learning for single-chip radar focuses on 2D outputs, we benchmark our transformer-based approach against two prior architectures for 2D BEV Occupancy prediction: a U-Net-based model (RadarHD~\cite{prabhakara2023high}), and a Swin Transformer-based model (T-FFTRadNet~\cite{giroux2023t}), with minor architecture modifications to conform to our data dimensions (App.~\ref{app:baseline_details}). We find that GRT-{\tt small} outperforms both baselines at all training splits (Fig.~\ref{fig:baselines}), demonstrating the suitability of transformer architectures for scalability.

\vspace{-1em}
\paragraph{Scaling Laws} Training 20 different models for 5 different sizes (Table~\ref{tab:sizes}) and dataset sizes ranging from 10-100\% of our data (Fig.~\ref{fig:scaling}), we observe a logarithmic improvement with data size of approximately 20\% improvement per 10$\times$ increase in data, similar to early observations in computer vision \cite{sun2017revisiting}. This can also be seen qualitatively (Fig.~\ref{fig:scaling_examples}), where models trained on more data produce much higher quality predictions. Similarly to vision transformers \cite{sun2017revisiting,zhai2022scaling}, we also observe that larger models are more data efficient, although the magnitude of difference that we observe is much smaller due to our limited dataset size.

\paragraph{Generalizability across different settings} We evaluate the ability of GRT to generalize across different settings by comparing a baseline model trained on combined \texttt{indoor}, \texttt{outdoor}, and \texttt{bike} data with models trained on each setting separately (Table~\ref{tab:ablations}). Despite the differences in these settings, the jointly trained model performs significantly better than models trained on each setting separately, confirming that data from different settings can be combined to train a single, stronger model.

\begin{table}
\caption{\textbf{Chamfer Distance (in meters) for 2D BEV occupancy prediction on the Coloradar dataset \cite{kramer2022coloradar} by location}; the geometric mean is listed to account for the varying difficulty of each location. A fine-tuned GRT model outperforms baselines trained only on Coloradar, including a state of the art diffusion-based model \cite{zhang2024towards} and a U-Net based model \cite{prabhakara2023high}.}
\vspace{-0.7em}
\centering
\footnotesize
\begin{tabular}{cccc}
\toprule
Trace & GRT (Ours) & Diffusion \cite{zhang2024towards} & RadarHD \cite{prabhakara2023high} \\
\toprule
\textit{Geometric Mean} & \textbf{0.98} & 1.19 & 1.73 \\
\hline
ARPG Lab & \textbf{0.78} & 0.96 & 1.73 \\
EC Hallways & \textbf{1.04} & \textbf{1.04} & 1.69 \\
Aspen & 0.61 & \textbf{0.51} & 0.91 \\
Longboard & \textbf{2.63} & 5.47 & 5.40 \\
Outdoors & \textbf{1.84} & 2.37 & 3.10 \\
Edgar & \textbf{0.36} & 0.44 & 0.60 \\
\toprule
\end{tabular}
\label{tab:coloradar}
\end{table}

\begin{figure*}
\centering
\includegraphics[width=0.93\textwidth]{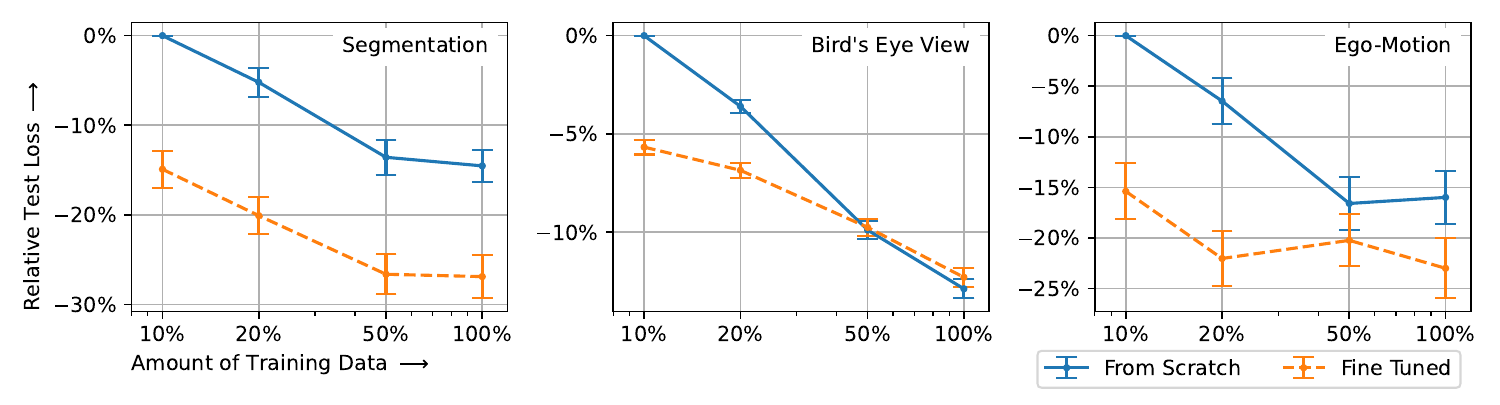}
\vspace{-1.3em}
\caption{\textbf{Fine tuning mmWave radar transformers to different downstream tasks.} Models and confidence intervals are measured relative to the \texttt{small} transformer trained (from scratch) on 10\% of the dataset. Pre-training and fine tuning strongly impacts data efficiency, equivalent to up to a $5\times$ increase in dataset size (observed in the Segmentation task).}
\label{fig:tuning}
\end{figure*}

\subsection{As a Base Model for Downstream Tasks}
\label{sec:tuning}

\paragraph{Dataset Fine-tuning} We fine-tuned a \texttt{small} GRT model on the Coloradar \cite{kramer2022coloradar} dataset using a BEV Occupancy objective, and benchmarked the resulting model against two prior approaches trained only on ColoRadar, including a state-of-the-art diffusion model, using the same data splits and evaluation procedure \cite{zhang2024towards}. Notably, despite using a modulation and resolution (128 range $\times$ 128 Doppler) which is not present in our dataset, GRT can be run without any architectural modifications, such as modifying the number of upsampling stages, as would be required for a convolutional architecture (App.~\ref{app:baseline_details}). After fine-tuning until validation loss convergece ($\approx$30 minutes of training using a single RTX 4090), GRT achieves substantially lower Chamfer distance than baselines trained only on ColoRadar, showing the value of easily tunable foundational models (Table~\ref{tab:coloradar}).

\vspace{-0.5em}
\paragraph{Task Fine-tuning} We also fine-tuned \texttt{GRT-small} for each of our secondary tasks using 10-100\% of our dataset, and compared the results with models trained from scratch on the same proportions of the dataset. Following this procedure, we find substantial performance gains equivalent to up to a 5$\times$ increase in dataset size compared to training from scratch (Fig.~\ref{fig:tuning}). This effect is especially pronounced when less data is available, with the performance benefits of fine tuning disappearing as the dataset is scaled for the BEV Occupancy objective but staying more or less constant for the Semantic Segmentation objective. We also observe this effect as a clear qualitative difference: fine-tuned models produce sharper and more accurate predictions than their counterparts trained from scratch (Fig.~\ref{fig:scaling_examples}).

\subsection{How Much More Data is Needed?}
\label{sec:how_much_data}

\begin{figure}
\centering
\includegraphics[width=0.9\columnwidth]{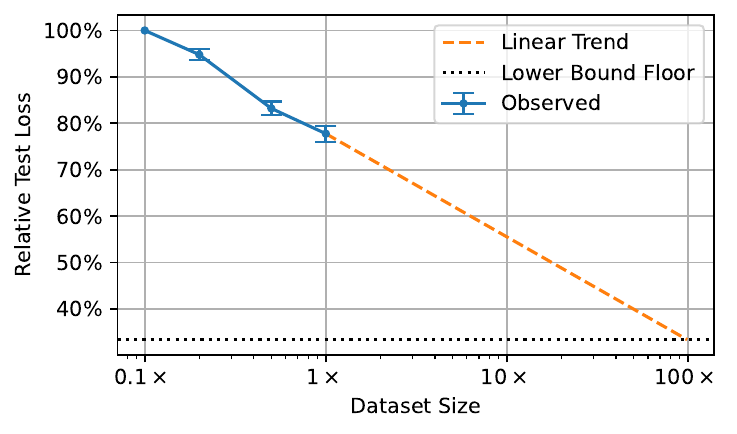}
\vspace{-1.2em}
\caption{A linear projection of the observed logarithmic scaling to a test loss lower bound suggests that logarithmic scaling cannot continue beyond 100$\times$ our current dataset}
\label{fig:linear_projection}
\end{figure}

Although we cannot directly observe performance saturation, we project how much data would be required to saturate a Radar Transformer using two different methods to arrive at a best guess of approximately 100M samples -- $100\times$ our current dataset.

\vspace{-0.5em}
\paragraph{Linear Projection of Scaling Laws} To lower-bound the possible test loss in our dataset, we trained a \texttt{small} model on the test set to approximate convergence. Assuming that the rate of improvement in model performance with increased training data cannot decrease, we extend our observed (Fig.~\ref{fig:scaling}) logarithmic scaling law to this lower bound to, in turn, estimate an upper bound for when the logarithmic trend will no longer hold (Fig.~\ref{fig:linear_projection}). This yields an estimate of $100\times$ our current dataset size. For additional details justifying our estimation of this bound, see App.~\ref{app:projections}.

\vspace{-0.5em}
\paragraph{Validation Curve Trends} We observe that GRTs tend to stop improving (with respect to validation loss) after $\approx$10 training epochs, regardless of model or dataset size (App.~\ref{app:projections}); this is similar to trends observed in the training of data-constrained LLMs, which are also observed to saturate around 10 epochs \cite{muennighoff2023scaling}. Using vision transformers, whose scaling laws are well studied \cite{zhai2022scaling} due to the availability of internet data, as a reference, we expect training saturation to occur around $10^2-10^4$M samples seen. Since each epoch in our dataset corresponds to $\approx$1M samples seen, this implies that $10\times$ to $1000\times$ our current dataset size is required to delay overfitting beyond this point.

\section{Conclusion}

In this paper, we train a Generalizable Radar Transformer using a large, 29 hour (1M sample) dataset collected using our open-source data collection system and demonstrate that our Radar Transformer can generalize across datasets and settings, can be readily fine-tuned, and exhibits logarithmic scaling.
While we believe that substantial gains are still possible through further data scaling, we hope that our dataset and baseline models will enable the community to revisit previous methods in new context and explore new capabilities made possible by a much larger dataset.


{
    \small
    \bibliographystyle{ieeenat_fullname}
    \bibliography{main}
}

\clearpage
\setcounter{page}{1}
\maketitlesupplementary
\appendix
\section{Data Collection and Dataset Details}
\label{app:dataset}

\begin{figure}
    \centering
    \includegraphics[width=\columnwidth]{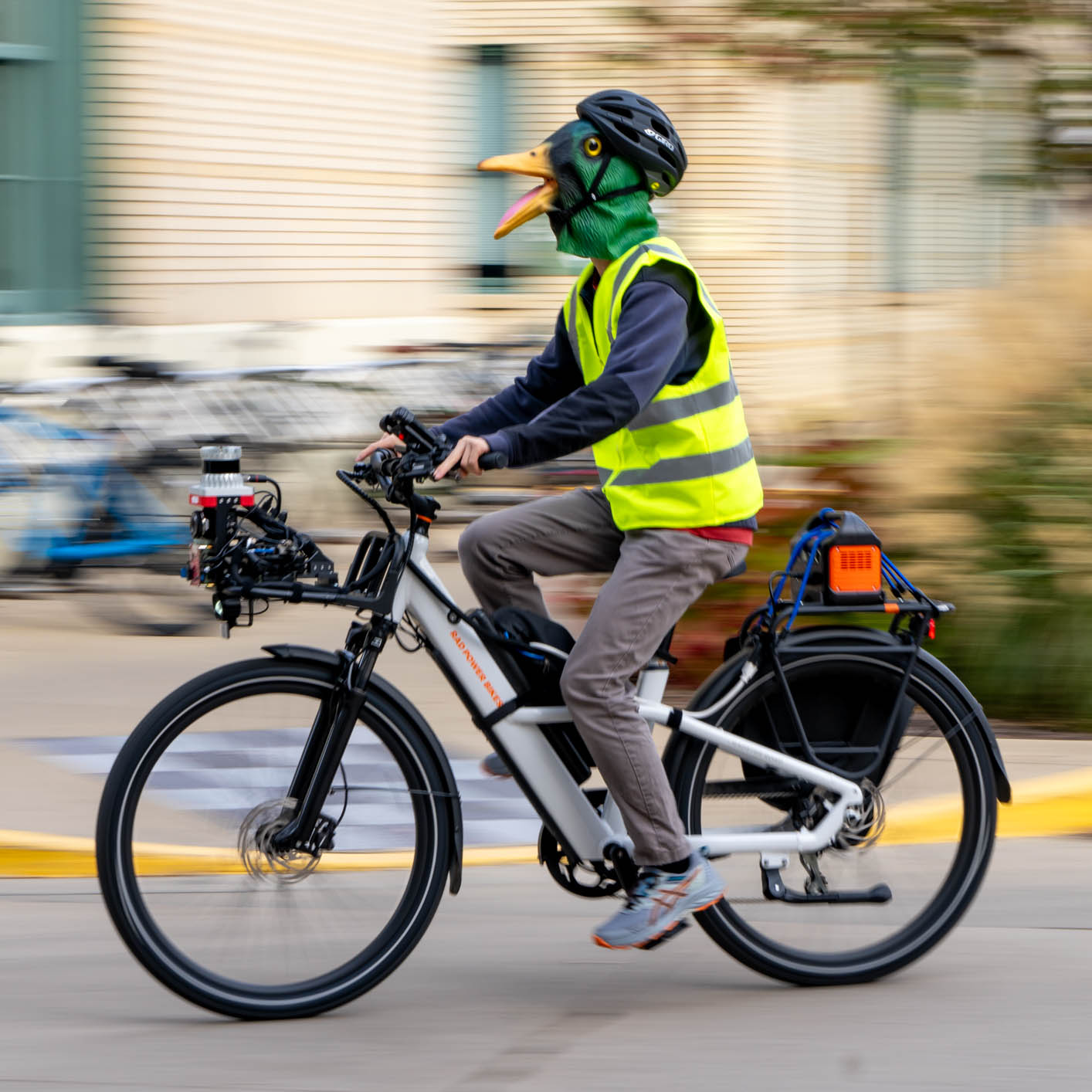}
    \vspace{-1.5em}
    \caption{\textbf{Our data collection system, \texttt{red-rover}, in its bike-mounted configuration.} The sensors are mounted to a front rack, while the support electronics are mounted in the center frame and the battery at the rear for balance and stability.}
    \label{fig:bike}
\end{figure}

\begin{figure*}
\begin{subfigure}[t]{\columnwidth}
    \centering
    \includegraphics[width=0.75\textwidth]{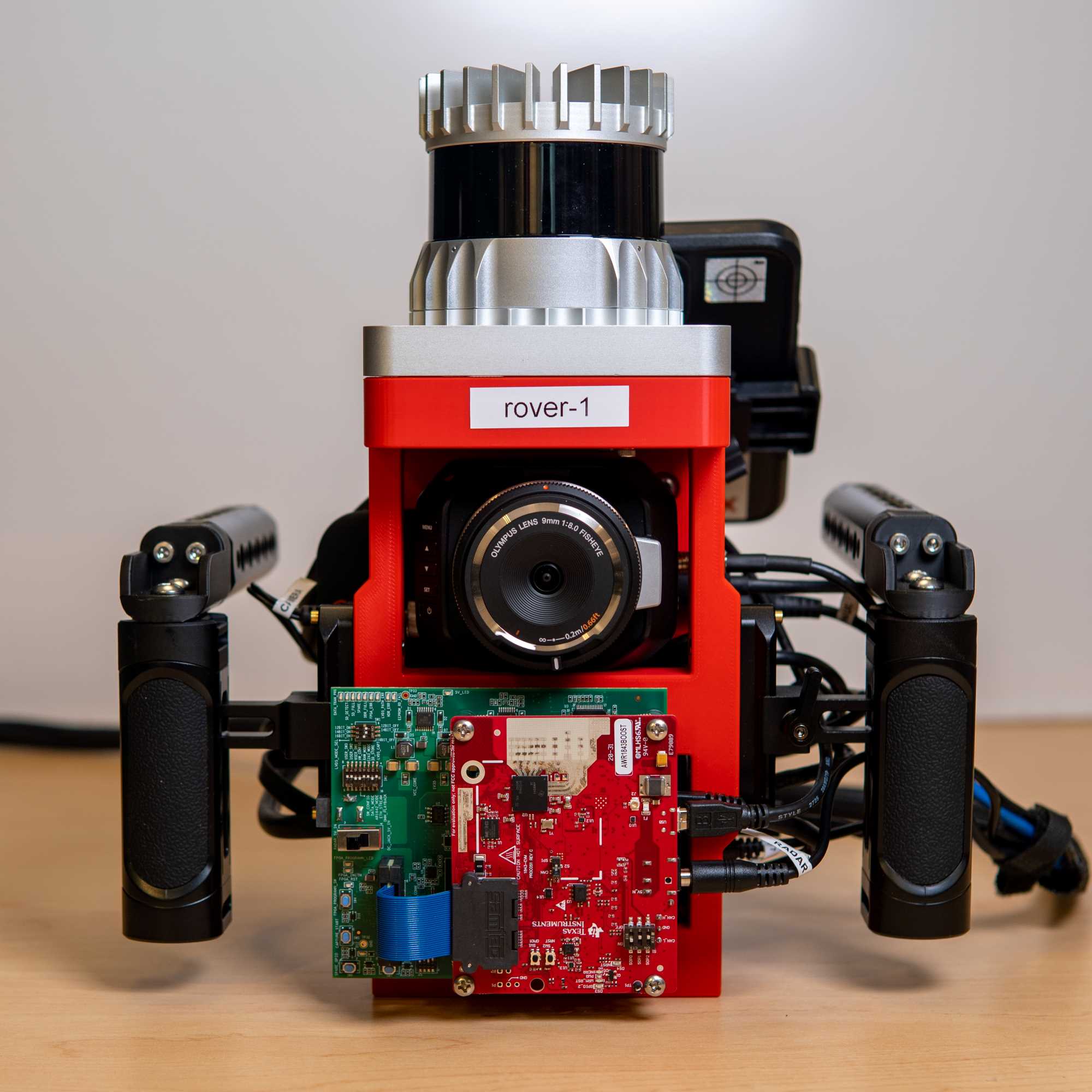}
    \caption{Data collection rig from the front; the Lidar, Camera, and Radar (red PCB) along with its capture card (green PCB) are visible, while the IMU is hidden inside the red (3D-printed) plastic structure.}
    \label{fig:rover_front}
\end{subfigure}
\hfill
\begin{subfigure}[t]{\columnwidth}
    \centering
    \includegraphics[width=0.75\textwidth]{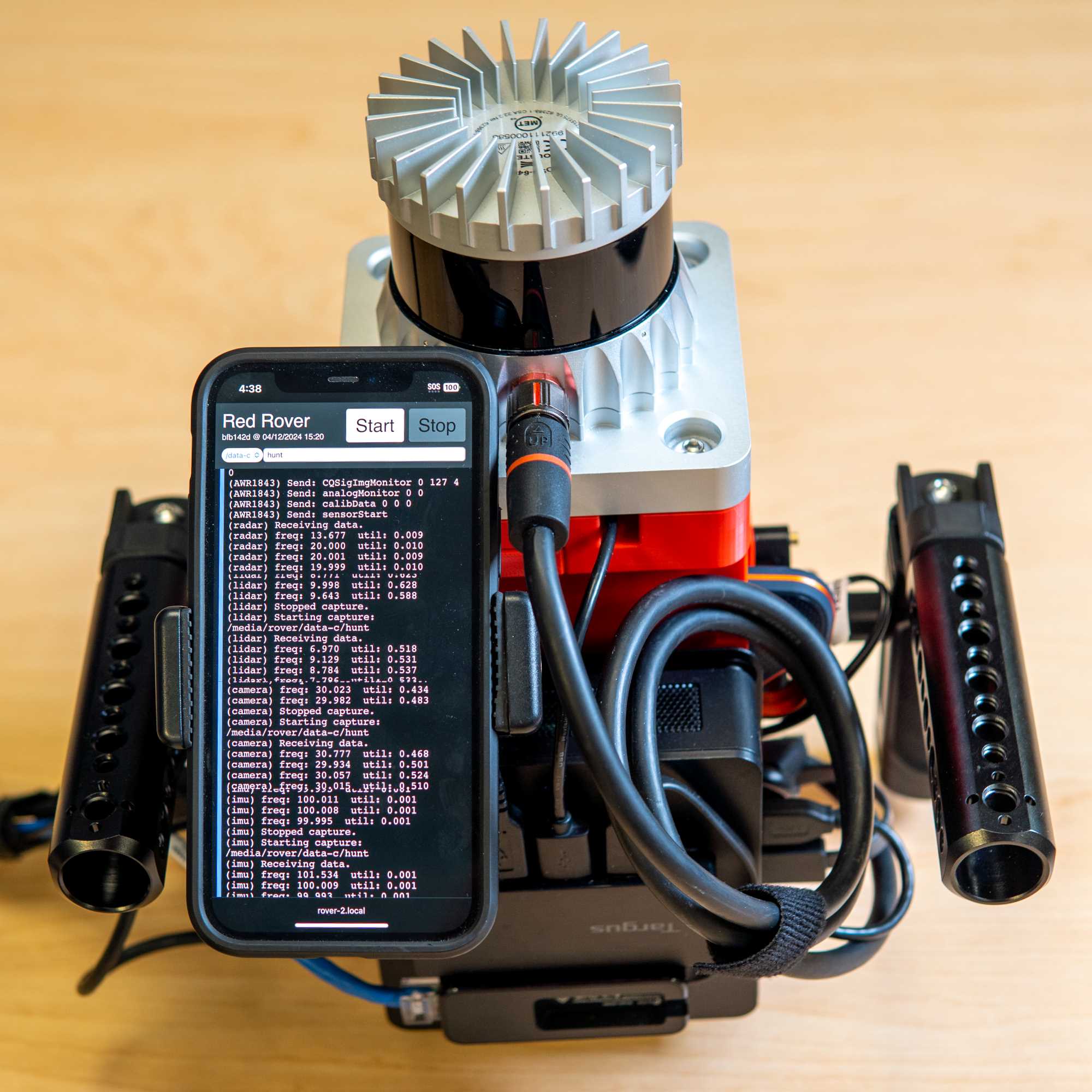}
    \caption{Data collection rig from behind, showing the control app; the app allows users to specify metadata, then \texttt{start} and \texttt{stop} data collection. A live console displays logged messages and errors for each sensor.}
    \label{fig:rover_back}
\end{subfigure}
\vspace{-0.3em}
\caption{\textbf{Close-up views of the handheld data collection rig from the front and back.}}
\label{fig:rover_closeup}
\end{figure*}

Our data collection system, \texttt{red-rover}\footnote{
As the successor to our previous \texttt{rover} data collection system \cite{huang2024dart}, \texttt{red-rover} is named for its distinctive red color. 
}, is fully open source. The data collection rig and control app are shown in detail in Fig.~\ref{fig:rover_closeup}, along with the system in its bike-mounted configuration in Fig.~\ref{fig:bike};
all parts used can be purchased either off-the-shelf or 3D-printed using an ordinary 3D printer.

\paragraph{Bill of Materials} The total cost of the bill of materials for our data collection system is \$4,440 for the base system (Table~\ref{tab:bom}). Note that the Lidar is the primary contributor to the cost of our data collection rig; while we use an OS0-128 (\$12,000), it can be substituted for an OS0-64 (\$8,000) or OS0-32 (\$4,000) without any hardware or software modifications, though at the cost of reduced Lidar data quality.

A detailed bill of materials including all parts used in the data collection rig (along with CAD files for 3D printed parts) is available in our \texttt{red-rover} project repository.

\paragraph{Resource Usage} For the configurations which we used to collect I/Q-1M, our data collection rig has the following overall characteristics:
\begin{itemize}
    \item Data rate: $\approx$120GB/hour ($\approx$33MB/s | 260Mbps), with some variation depending on the compressibility of the data. In practice, we do not find storage to be a substantial limitation, with the total dataset size being $\approx$3.5TB.
    \item Power consumption: $\approx$80W average. Using a 240Wh AC battery bank, this results in around 3 hours of battery life.
\end{itemize}

\subsection{Sensors}
\label{app:sensors}

Our data collection rig includes a radar, lidar, camera, and IMU, and records a total data bitrate of $\approx$260~Mbps. Almost half of the bitrate is consumed by the radar (126~MBbps), with the remainder being split between the Camera and Lidar, with a negligible amount data recorded from the IMU.

\begin{table}
\caption{\textbf{Bill of materials and approximate cost of major components} as of time of writing, in US Dollars; carrying equipment (e.g., backpack, E-bike) and miscellaneous items under \$100 (e.g. cables, screws) are not listed.}
\vspace{-0.7em}
\centering
\small
\begin{tabular}{cc}
\toprule
Item & Cost \\
\toprule
Ouster OS0-32/64/128 Lidar & \$4,000-\$12,000 \\
Data Collection Computer & \$1,000 \\
Black Magic Micro Studio Camera & \$1,000 \\
Magewell USB-SDI Capture Card & \$300 \\
OM Systems 9mm f/8 Fisheye Lens & \$100 \\
TI DCA1000EVM Radar Capture Card & \$600 \\
TI AWR1843Boost Radar & \$300 \\
XSens MTi-3 AHRS Development Kit & \$450 \\
\hline
External Storage Drive & \$330 \\
Battery & \$240 \\
Hardware for Handles & \$120 \\
\hline
\textbf{Total} & \$4,440 + Lidar \\
\toprule
\end{tabular}
\label{tab:bom}
\end{table}

\begin{figure}
    \centering
    \vspace{-0.5em}
    \includegraphics[width=\columnwidth]{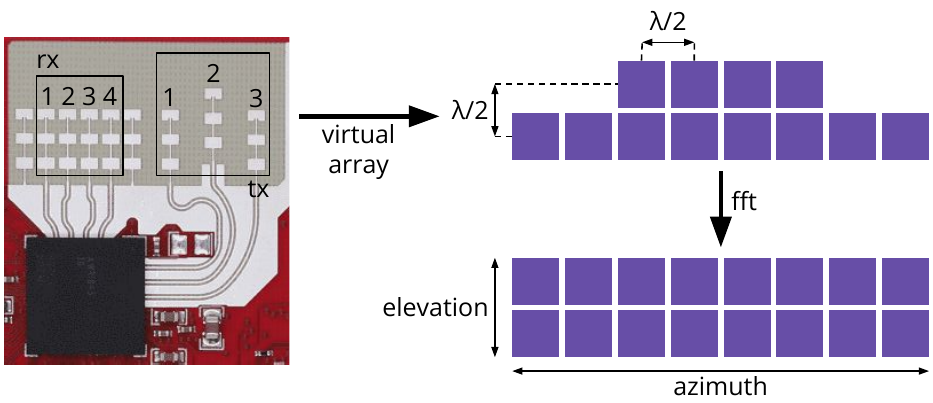}
    \vspace{-1.7em}
    \caption{\textbf{Antenna configuration of the TI AWR1843Boost radar}. The 12 virtual antennas (top right) created by the radar's 3TX $\times$ 4RX antenna array (left) result in 2 elevation and 8 azimuth bins (lower right).}
    \label{fig:antenna-array}
\end{figure}

\paragraph{Radar} ``Boost'' development boards for the TI 77GHz single-chip mmWave radar family\footnote{
    TI produces ``Boost'' series development boards across its range of 77GHz radars including the AWR1843Boost, the largely identical IWR1843Boost, and the AWR1642Boost, which is equivalent to the AWR1843Boost with its middle transmit antenna removed.
} are commonly used in academic research, and we are not aware of any raw single-chip radar datasets -- or tooling for data collection -- which uses other radars. As such, we use the AWR1843Boost Radar (and a DCA1000EVM capture card), which is commonly used in prior literature \cite{lim2021radical,prabhakara2023high,huang2024dart}.

The AWR1843Boost has 3 transmit (TX) and 4 receive (RX) antennas, resulting in 8 azimuth and 2 elevation bins (Fig.~\ref{fig:antenna-array}). We configured our radar to record 256 range bins and 64 Doppler bins at 20~Hz, with varying range and Doppler resolutions depending on the setting; see Table~\ref{tab:configurations} for detailed specifications. In our dataset, we also collect raw, uncompressed I/Q streams which are quantized as 16-bit integers by the radar; with the modulations used in our dataset, the radar has a total bitrate of 126~Mbps.

Crucially, we do not to use a high-resolution imaging radar (e.g., the $12\times 16$ antenna TI MMWCAS-RF-EVM): in addition to their larger size, weight, and power, imaging radars have an order-of-magnitude higher raw data rate (e.g., $\approx$2gbps for an equivalent modulation using the TI MMWCAS-RF-EVM), which substantially increases the engineering and infrastructure cost of collecting raw data, while also making continuous live streaming and real-time deployment impractical.

\begin{table}
\caption{\textbf{Full radar configurations for each setting.} With a fixed frame size of $N_r = 256$ samples/chirp and $N_d = 64$ chirps/frame, configuring the radar's chirp rate, ADC sample rate, and chirp slope determines the range resolution $\Delta R = \frac{F_s c}{2N_r}$ and Doppler resolution $\Delta D = 
\frac{\lambda}{2N_dT_c}$ along with maximum range $R_{\text{max}} = \frac{F_S c}{2s}$ and maximum Doppler $D_{\text{max}} = \frac{\lambda}{4T_c}$, where $\lambda$ is the radar's wavelength (77GHz; $\lambda=3.9$mm) and $c$ is the speed of light.}
\vspace{-0.7em}
\centering
\footnotesize
\begin{tabular}{cccc}
\toprule
Setting & \texttt{indoor} & \texttt{outdoor} & \texttt{bike} \\
\toprule
Chirp Time $T_c$ & 777$\mu$s & 537$\mu$s & 120$\mu$s \\
Sample Rate $F_s$ & 5MHz & 5MHz & 10MHz \\
Chirp Slope $S$ & 67MHz/$\mu$s & 34MHz/$\mu$s & 34MHz/$\mu$s \\
$\Delta R$ & 4.4cm & 8.7cm & 8.7cm \\
$R_\text{max}$ & 11.2m & 22.4m & 22.4m \\
$\Delta D$ & 3.8cm/s & 5.6cm/s & 24.9cm/s \\
$D_\text{max}$ & 1.2m/s & 1.8m/s & 8.0m/s \\
\toprule
\end{tabular}
\label{tab:configurations}
\end{table}

\paragraph{Lidar} We use an Ouster OS0-128 recording 2048 azimuth bins (1024 forward-facing) and 128 elevation beams at 10Hz. In practice, we find that the OS0 Lidar has a maximum range of 20-25m: while points further away can still \textit{sometimes} be detected, objects are not consistently detected. As such, while our radar can detect objects at much further ranges, the Lidar forces us to restrict the maximum range used in our dataset; we plan to acquire a longer-range Lidar for future iterations of our dataset.

The lidar depth is LZMA-compressed, resulting in a bitrate of 14~Mbps. While we do not use these channels in our paper, we also collect the reflectance and near infrared background intensity, with a typical data rate of 9~Mbps and 22~Mbps, respectively.

\begin{table*}
\caption{\textbf{Comparison with other mmWave radar datasets with raw data (4D data cube or equivalent), and a selection of other large datasets without raw data.} A \textit{frame} in our table refers to the number of unique radar-sensor tuples. Our dataset is significantly larger than previous radar datasets, enabling us to scale up training.}
\vspace{-0.7em}
\centering
\footnotesize
\begin{tabular}{cccccc}
\toprule
Radar Type & Dataset & 4D Data Cube & Dataset Size & Other Sensors \\
\toprule
\multirow{9}*{Single Chip} & \textbf{IQ-1M (Ours)} & \textbf{Yes} & \textbf{29 hours (1M frames)} & {Lidar}, Camera, IMU \\
& MilliPoint \cite{cui2024milipoint} & No (3D Points) & 6.3 Hours (545k frames) & Depth Camera \\
& nuScenes \cite{caesar2020nuscenes} & No (3D Points) & 5.5 hours (400k frames) & {Lidar}, Camera, IMU, GPS \\
& RaDICal \cite{lim2021radical} & \textbf{Yes} & 3.6 Hours (394k frames) & Depth Camera, IMU \\
& Coloradar \cite{kramer2022coloradar} & \textbf{Yes} & 2.4 hours (82k frames) & {Lidar}, IMU \\
& CRUW \cite{wang2021rodnet} & No (2D Map) & 3 hours (400k frames) & Stereo Cameras \\
& RadarHD \cite{prabhakara2023high} & \textbf{Yes} & 200k frames & Lidar \\
& CARRADA \cite{ouaknine2021carrada} & No (3D Cube) & 21 Minutes (13k frames) & Camera \\
& RADDet \cite{zhang2021raddet} & \textbf{Yes} & 10k frames & Camera \\
\hline
\multirow{3}*{Cascaded} & Radatron \cite{madani2022radatron} & No (3D Cube) & 4.2 hours (152k frames) & Camera \\
& K-radar \cite{paek2022k} & \textbf{Yes} & 35k frames & {Lidar}, Camera, IMU, GPS \\
& RADIal \cite{rebut2022raw} & \textbf{Yes} & 20k frames & {Lidar}, Camera, GPS \\
\hline
\multirow{3}*{Mechanical} & Oxford Radar RobotCar \cite{barnes2020oxford} & No (2D Image) & 17 Hours (240k Frames) & {Lidar}, Camera, GPS \\
& RADIATE \cite{sheeny2021radiate} & No (2D Image) & 5.0 hours (72k frames) & {Lidar}, Camera, GPS \\
& Boreas \cite{burnett2023boreas} & No (2D Image) & 350km & Lidar, Camera \\
\toprule
\end{tabular}
\label{tab:other_datasets}
\end{table*}

\paragraph{Camera} We use a Black Magic Micro Studio Camera with an OM Systems 9mm f/8 Fisheye lens, recording at 1080p, 30~fps; frames are recorded as a MJPEG video, resulting in a typical bitrate of 88~Mbps. To minimize motion blur, the camera is set to 18~db gain; the shutter speed is set to automatic. Note that while our camera and capture card are capable of 60~fps recording, we record only 30~fps since we find that 30~fps recording is far more stable than 60~fps (especially with regard to dropped frames), and since since the downstream tasks which we envision cannot easily take advantage of 60~fps video.

\paragraph{IMU} We include a XSens MTi-3 IMU which is used for Cartographer SLAM in conjunction with our Lidar. The IMU records acceleration, angular velocity, and rotation at 100~Hz, with a total bitrate of 35~Kbps.

\subsection{Time Synchronization}

While each sensor is recorded against the same system clock, we asynchronously record each at its full ``native'' speed. To generate radar-lidar-camera samples, we align higher frequency sensors (radar -- 20Hz; camera -- 30Hz) to the Lidar (10Hz) by selecting the nearest sample in time to each lidar frame. Since our data collection implementations for each sensor have variable initialization and de-initialization time, we also trim regions at the start and end of each trace which do not have coverage from all sensors.

\subsection{Comparison with Other Datasets}
\label{app:other_datasets}

Table~\ref{tab:other_datasets} enumerates a number of radar datasets sorted by radar type, along with their sizes and included sensors. Since different types of radars have substantially different operating modes, modulations, and data characteristics and dimensions, we focus on single-chip radars. In this category, prior datasets generally use TI single-chip radars such as the TI AWR1843 family which we use \cite{ti:AWR1843AOP}.

\paragraph{Fine-Tuning Experiment} In our fine-tuning experiments, we elected to use the Coloradar \cite{kramer2022coloradar} dataset due to its inclusion of high-quality Lidar depth data and extensive prior work using this dataset. We considered, but opted not to use, the following datasets:
\begin{itemize}
    \item RaDICal \cite{lim2021radical}: the depth cameras used have poor performance, especially beyond very close ranges. Likely because of this issue, we are also not aware of a substantial body of prior work using this dataset as a benchmark.
    \item RADDet \cite{zhang2021raddet}: RADDet is an extremely popular object detection dataset, and a good candidate for fine-tuning. Unfortunately, while we have been able to obtain the raw ADC data, the ground truth labels, video, and other metadata are no longer available as of time of writing.
    \item CRUW \cite{wang2021rodnet}: While the original CRUW dataset appears to include lower-level data, only 2D range-azimuth maps are publicly available.
\end{itemize}

\subsection{Dataset Details}
\label{app:settings}

\begin{figure*}
\centering
\begin{subfigure}[t]{0.49\textwidth}
    \centering
    \includegraphics[width=\textwidth]{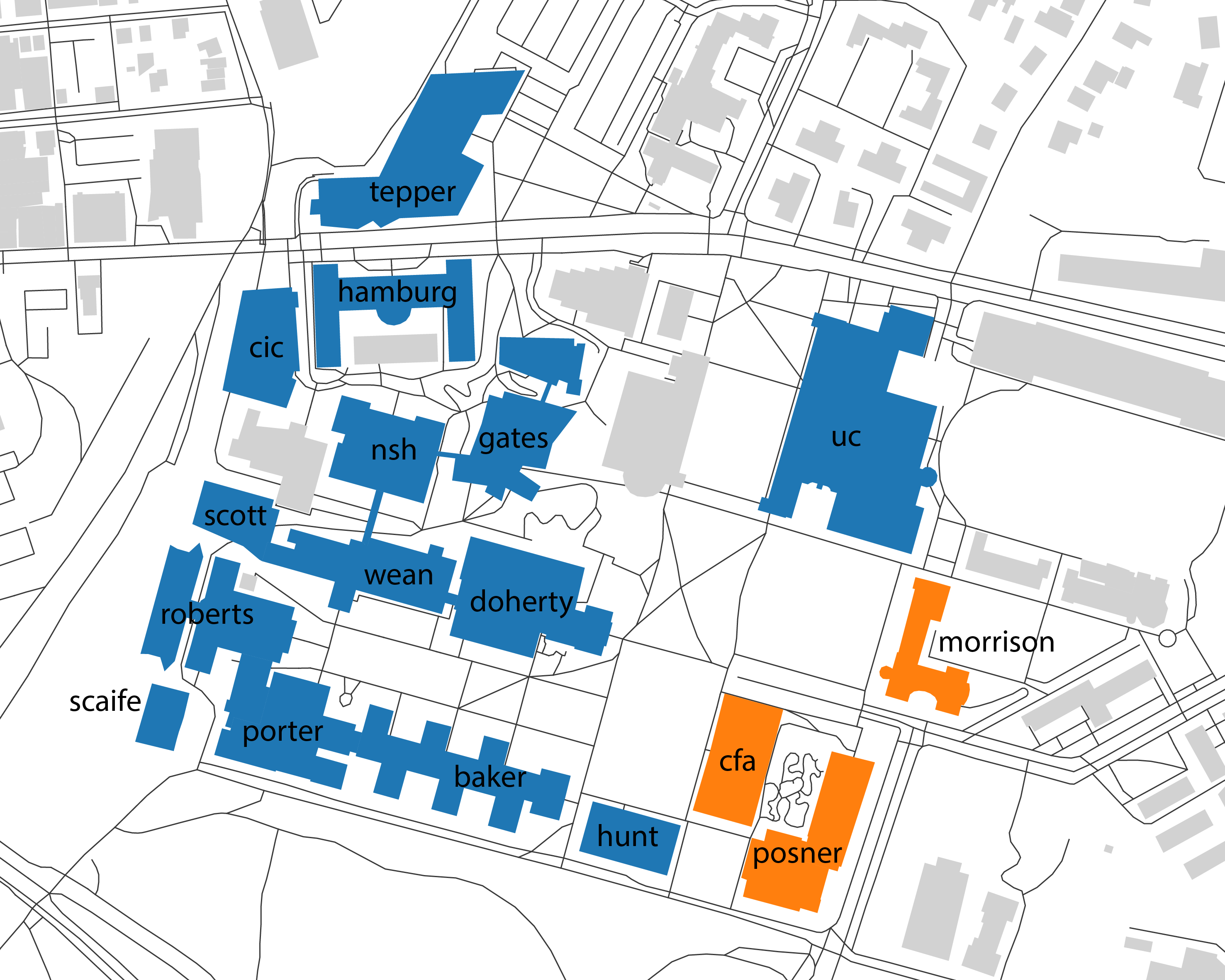}
    \caption{Buildings visited in the \texttt{indoor} split.}
    \label{fig:map_indoor}
\end{subfigure}
\begin{subfigure}[t]{0.49\textwidth}
    \centering
    \includegraphics[width=\textwidth]{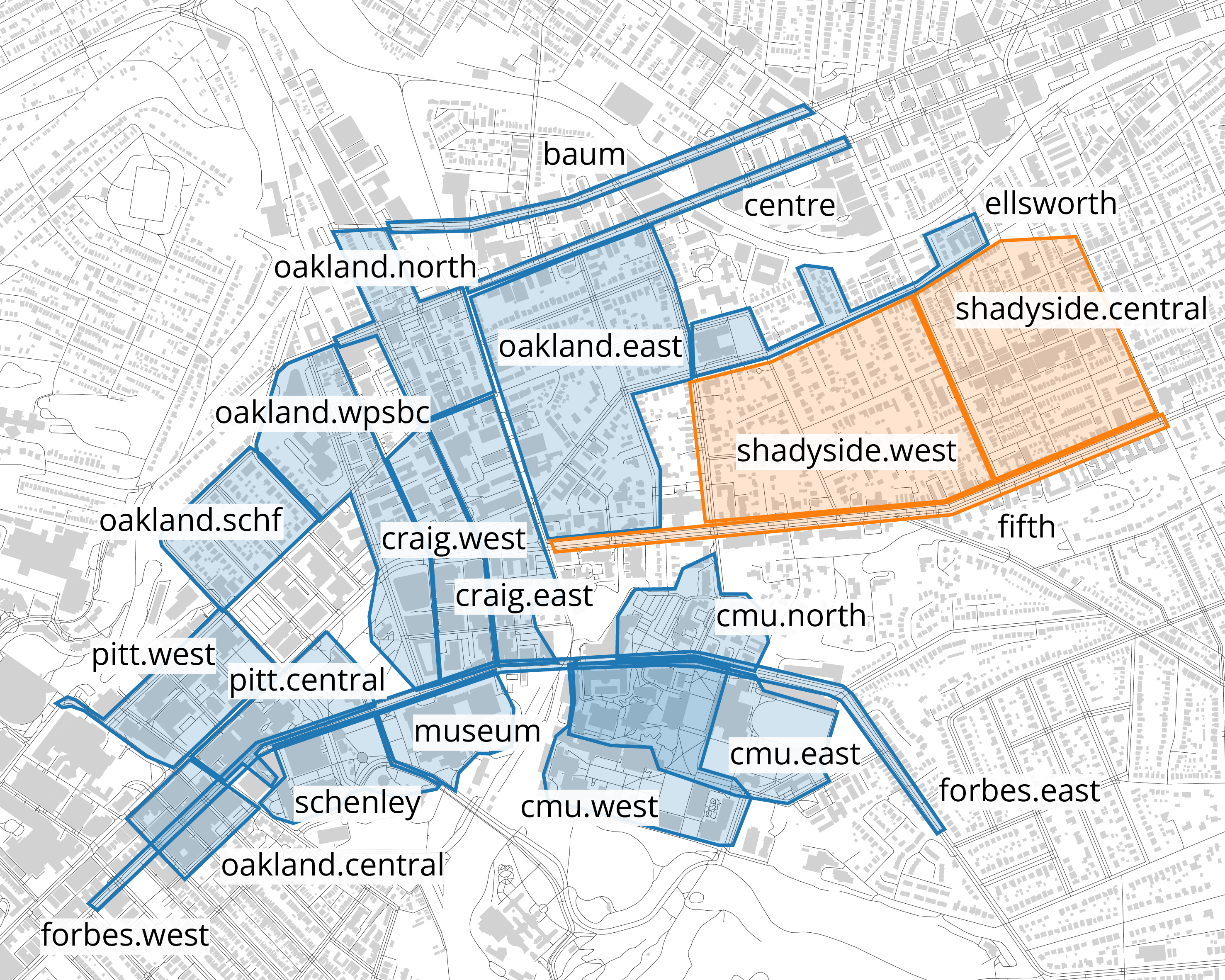}
    \caption{Areas covered by traces in the \texttt{outdoor} split.}
    \label{fig:map_outdoor}
\end{subfigure} \\
\vspace{0.4em}

\begin{subfigure}[t]{0.49\textwidth}
    \centering
    \includegraphics[width=\textwidth]{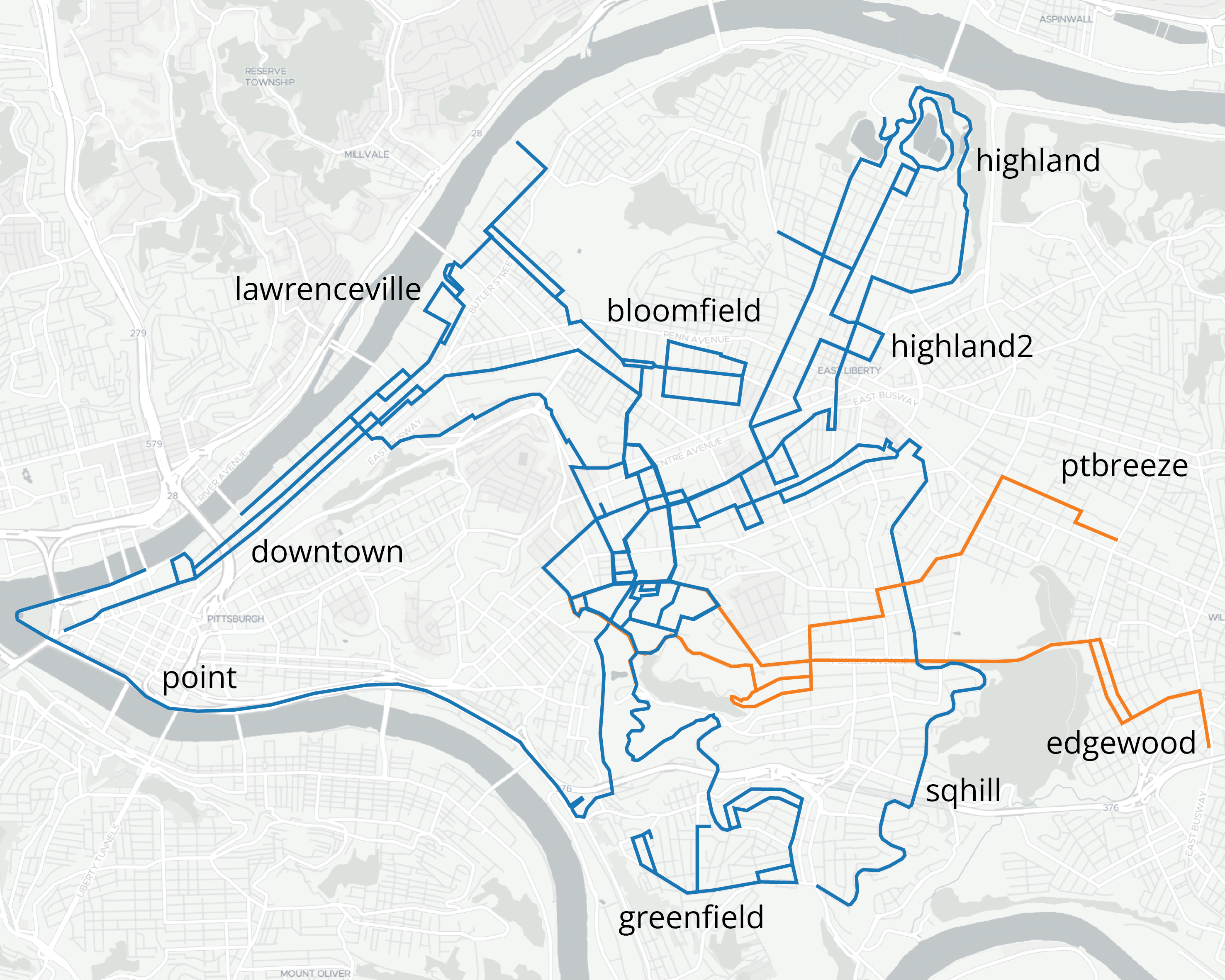}
    \caption{Traces in the \texttt{bike} split.}
    \label{fig:map_bike}
\end{subfigure}
\begin{subfigure}[t]{0.49\textwidth}
    \centering
    \includegraphics[width=\textwidth]{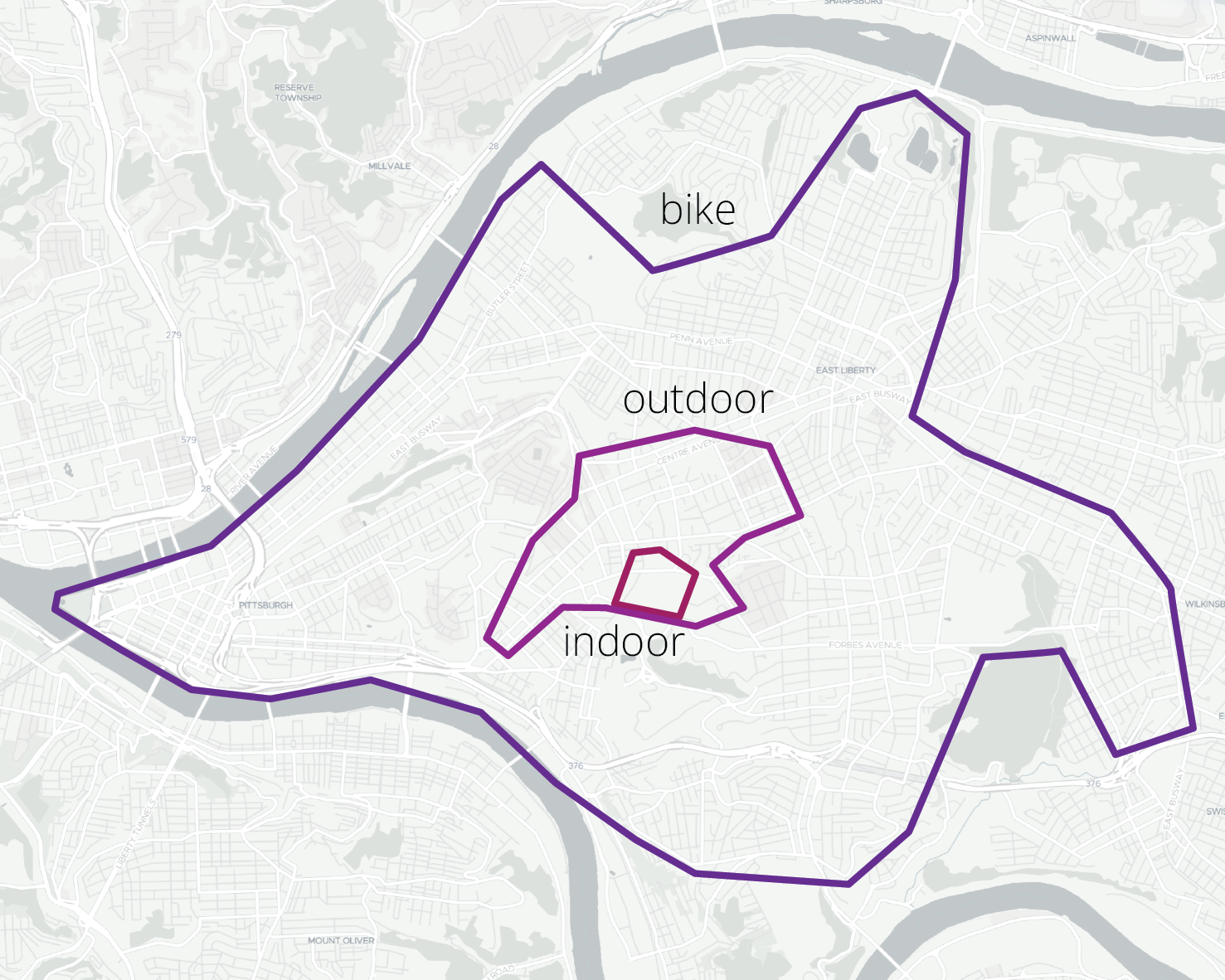}
    \caption{Overview of the data collection extents for each setting.}
    \label{fig:map_overview}
\end{subfigure}

\vspace{-0.2em}
\caption{\textbf{Maps of the train (blue) and test (orange) splits for each setting.}}
\label{fig:maps}
\end{figure*}

\begin{figure*}
\includegraphics[width=\textwidth]{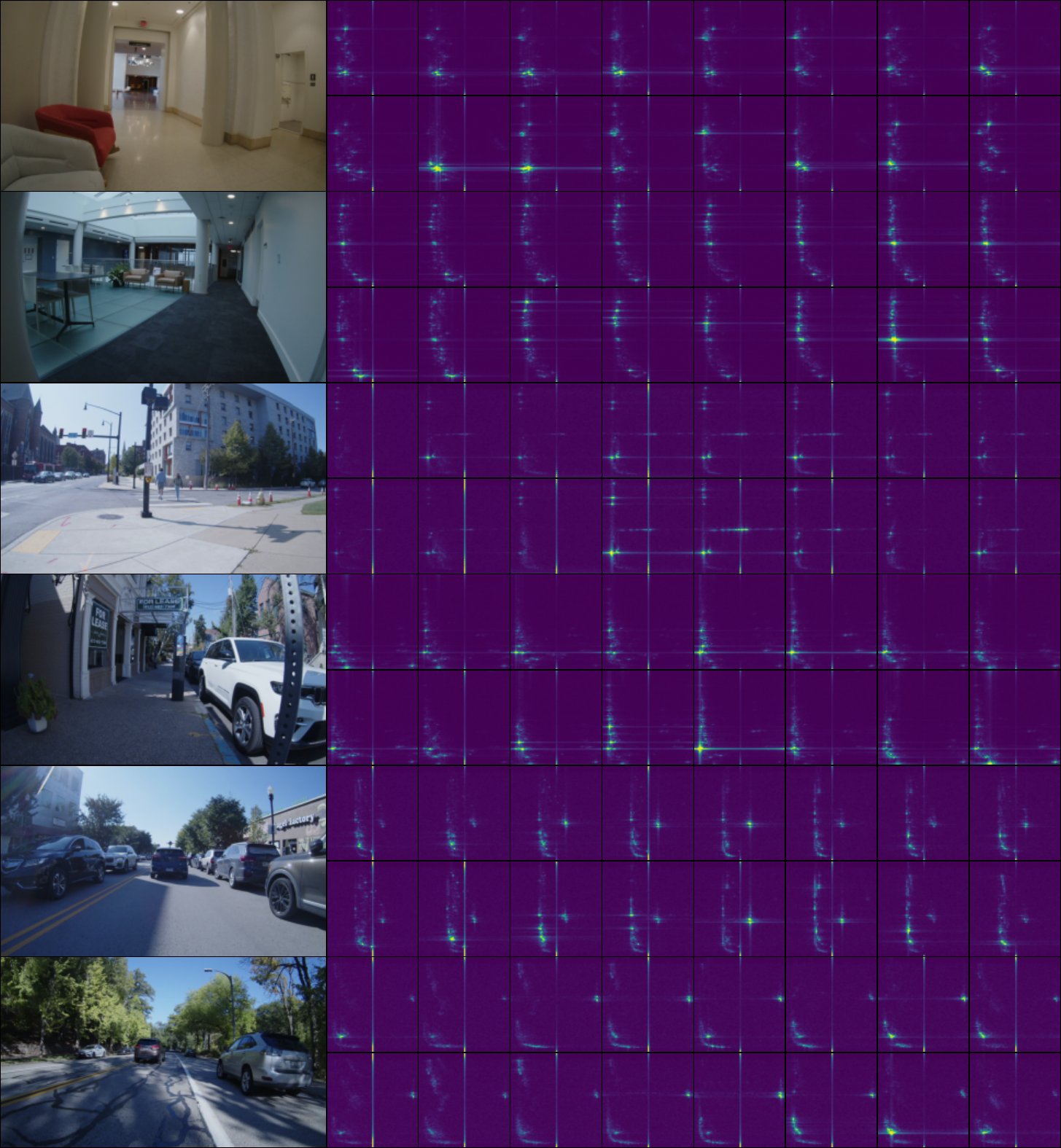}
\caption{\textbf{Representative samples from our dataset showing camera and range-Doppler frames from the \texttt{indoor} (top), \texttt{outdoor} (middle), and \texttt{bike} (bottom) settings}. Each radar plot shows the range-Doppler image of a single (azimuth, elevation) bin. When the radar is configured with a range and Doppler resolution which is appropriate for each setting, the resulting range-Doppler frames are remarkably similar at a visual level. Note that common types of radar noise and artifacts such as a zero doppler artifact (the straight line at the center of each frame) and range-Doppler bleed (horizontal and vertical lines coming from bright reflectors) are clearly visible in these examples.}
\label{fig:sample_data}
\end{figure*}

Our dataset was collected on and around the CMU campus and in the Pittsburgh area, and includes three data settings: handheld indoors, handheld outdoors, and on a bike. In addition to maps of data collection areas (Fig.~\ref{fig:maps}), we also provide representative samples from each dataset (Fig.~\ref{fig:sample_data}).

\vspace{0.5em}
\noindent\texttt{indoor}:
The indoor setting was collected from publicly accessible areas within the CMU campus. Each trace generally represents a different floor (or multiple floors, in cases where each floor is relatively small). Additionally, each floor was covered twice: once in a forward-facing configuration (with the velocity mostly aligned with the radar's orientation), and once in a ``sideways facing'' configuration (with the velocity mostly orthogonal to the radar).

\vspace{0.5em}
\noindent\texttt{outdoor}:
Roughly 30-minute-long traces were collected within contiguous areas with minimal overlap, with the radar generally facing forward. The areas visited include CMU and Pitt university campuses, commercial areas ranging from medium to high density, and residential areas ranging from single-family detached to high rise apartment buildings, as well as a variety of streets ranging from small alleys to busy ``stroads.''

\vspace{0.5em}
\noindent\texttt{bike}:
Data was collected on approximately 60-minute-long round-trips\footnote{
As one the authors was struck by a vehicle while collecting data on bike, we urge any efforts to replicate or extend this split to minimize mental load during data collection and use caution when planning routes. Thankfully, the author was uninjured, though the radar was destroyed.
} originating from our lab; each trace is split into an inbound and outbound leg covering mostly the same path, but in different directions. Note that there is some overlap between the areas covered in the train and test splits at ``bottlenecks'' near the CMU campus; when viewed as a whole, we believe this is negligible.



\section{Method Details}
\label{app:method}

GRT uses a standard encoder-decoder transformer network (App.~\ref{app:hyperparameters}). We also document our data augmentations (App.~\ref{app:augmentations}), training tasks (App.~\ref{app:tasks}), evaluation procedure (App.~\ref{app:procedure}), and baselines (App.~\ref{app:baseline_details}).

\subsection{GRT Training \& Hyperparameters}
\label{app:hyperparameters}

\begin{table}
\centering
\caption{\textbf{Key Hyperparameters for GRT.} Except for model layers and dimensionality, which we perform scaling law ablations on, these hyperparameters are taken from common transformer design practices as of time of writing.}
\small
\begin{tabular}{cc}
\toprule
Input Patch Size & 128 \ ($4 \times 2 \times 8 \times 2$) \\
Input Number of Patches & 2048 \\
Output Number of Patches & 1024 \\
\hline
Layers & 4 -- 18 \\
Model Dimension & 256 -- 768 \\
Dimensions Per Head & 64 \\
Expansion Ratio & 4.0 \\
\hline
Transformer Norm & ``pre-norm'' \\
Activation & GeLU \\
Dropout & 0.1 \\
\hline
Batch Size & 32 \\
Optimizer & AdamW \\
Warmup & 100 Steps \\
Learning Rate & $10^{-4}$ \\
\toprule
\end{tabular}
\label{tab:hyperparameters}
\end{table}

GRT uses a standard transformer architecture with sinusoidal positional encodings, and experimentally obtained radar-specific patching parameters. For a summary of key hyperparameters and architecture parameters, see Table~\ref{tab:hyperparameters}.

\paragraph{Architecture} Unless specified otherwise, GRT's architecture uses the following common parameters:
\begin{itemize}
    \item We always use a simple linear patch and unpatch layers, with the appropriate output dimensionality depending on the task.
    \item All layers use a GeLU activation \cite{hendrycks2016gaussian}.
    \item Transformers use ``pre-norm'' (norm before, instead of after the transformer layer), which is generally regarded as more stable \cite{xiong2020layer}; when using ``post-norm'' (as in the original transformer architecture \cite{vaswani2017attention}), we find that GRT often diverges due to numerical instability at initialization.
    \item Each transformer layer has a fixed expansion ratio of 4.0 and a dropout of 0.1.
\end{itemize}

\paragraph{Positional Encodings} In both encoder and decoder positional embeddings, we use a simple N-dimensional encoding which divides the number of features equally between each dimension and independently applies a sinusoidal encoding for the coordinate in that axis. To facilitate fine-tuning for tasks with different output resolutions, we also normalize the frequencies by the total length of each axis so that different resolutions result in the same frequency range.

\paragraph{Training} When training GRT, we use the following parameters for all models:
\begin{itemize}
    \item We apply a range of data augmentations, which we find to provide a $\approx 5\%$ benefit (App.~\ref{app:augmentations}).
    \item We always use a fixed batch size of 32. When training on platforms with different GPU counts, the batch is split equally between each GPU.
    \item All models are trained with a learning rate of $10^{-4}$ using the AdamW \cite{loshchilov2017fixing} optimizer with a warmup period of 100 steps. We find this warmup period to be essential in order to avoid initialization instability and NaN gradients.
    \item  Each model was trained until the validation loss stopped decreasing, as defined by three consecutive checkpoints without a decrease in validation loss, with two checkpoints taken each epoch.
\end{itemize}

\paragraph{Fine-Tuning} Fine-tuning uses the same procedure as for training, including termination after the validation loss stops decreasing. The model is not frozen, with all weights being trainable during the tuning process. In cases where the output dimension does not match the input dimension (e.g., 8-channel one-hot classification outputs for the Semantic Segmentation objective vs. 1-channel binary classification outputs for occupancy objectives), the output layer is also re-initialized.

\paragraph{Patch Size} We use a patch size of 128 bins (4 range, 2 Doppler, 8 azimuth, 2 elevation) in the encoder, resulting in 2048 input patches, and square (or cubic) patches for each output sized to maintain a fixed decoder sequence length of 1024 patches. Note that this ``patches out'' the azimuth and elevation axes in the encoder; while we empirically determined that this leads to the best performance on our dataset (Sec.~\ref{sec:ablations}), we expect the optimal patch dimensions to vary depending on the input radar resolution.

\paragraph{Patch Size Alternatives} In addition to the above patch size, we tested the following alternatives:
\begin{itemize}
    \item \textbf{Range-Doppler-azimuth-elevation}: carefully selecting our patch size to keep all four axes, we create 16 range $\times$ 8 Doppler $\times$ 8 azimuth $\times$ 2 elevation patches. This results in a $4.18\pm 1.09\%$ increase in test loss. 
    \item \textbf{Range-azimuth-elevation}: we eliminate the Doppler axis for 128 Range $\times$ 8 azimuth $\times$ 2 elevation patches. This results in a $6.27 \pm 1.10\%$ increase in test loss.
    \item \textbf{Doppler-azimuth-elevation}: we eliminate the Range axis (as much as possible) to obtain 64 Doppler $\times$ 2 range $\times$ 8 azimuth $\times$ 2 elevation patches. This results in a $6.22 \pm 1.11\%$ increase in test loss.
\end{itemize}

\subsection{Data Augmentations}
\label{app:augmentations}

We develop a range of data augmentations, which we ablate in two groups: \textit{Scale, Phase, and Flip Only}, and \textit{Full} augmentations, which we use by default.

\paragraph{Scale, Phase, and Flip Only} This group includes ``simple'' augmentations which can be calculated pixel-wise:
\begin{itemize}
    \item \texttt{radar\_scale}: random scaling applied to the magnitude of the 4D radar data cube, with log-normal distribution $\exp(\mathcal{N}(0, 0.2^2))$ clipped to $[\exp(-2), \exp(2)]$.
    \item \texttt{radar\_phase}: random phase offset of $\text{Unif}(-\pi, \pi)$ applied to the phase of the 4D radar data cube (except in ablations where no phase information is provided to the model).
    \item \texttt{azimuth\_flip}: random flipping (with probability 0.5) along the azimuth axis, i.e., swapping left and right. Note that this augmentation also affects the ground truth for each task.
    \item \texttt{doppler\_flip}: random flipping (with probability 0.5) along the Doppler axis, i.e. swapping positive and negative Doppler. This is equivalent to reversing the direction of travel of the sensor and all other objects in the scene; as such, this augmentation also affects the ground truth velocity by reversing the velocity vector.
\end{itemize}
Note that we do not apply an \texttt{elevation\_flip} since the ground is always down!

\paragraph{Full} In addition to the above augmentations, we include augmentations which are equivalent to random cropping:
\begin{itemize}
    \item \texttt{range\_scale}: ranges are multiplied by a $\text{Unif}(1.0, 2.0)$ scale, with the radar data cube being cropped and scaled appropriately using a bilinear interpolation. The ground truth occupancy and Bird's Eye View are also scaled accordingly.
    \item \texttt{speed\_scale}: Doppler velocities are multiplied by a log-normal $\exp(\mathcal{N}(0, 0.2^2))$ distribution, clipped to $[\exp(-2), \exp(2)]$. All scaling is done with bilinear interpolation. If the velocity is scaled down, we zero-fill any extra bins; if velocity is scaled up, we ``wrap'' the Doppler velocity around to emulate the ambiguity of Doppler velocity modulo the Doppler bandwidth. Finally, the ground truth ego-motion velocity is scaled to match.
\end{itemize}

\subsection{Task Details}
\label{app:tasks}

\paragraph{3D Occupancy Classification} Our 3D polar occupancy task uses a binary cross-entropy loss on polar grid cells which are created by the product set of the radar's range resolution with the Lidar's azimuth and elevation resolution. The loss is further scaled and balanced for the following:
\begin{itemize}
\item To facilitate joint training between different radar modulation, we normalize distances by the radar range resolution, resulting in a fixed output grid for each setting.
\item To correct for the sparsity of 3D occupancy grids, occupied cells are weighted greater (64.0) than unoccupied cells (1.0).
\item Since polar occupancy cells are larger when further away, we correct for the cell size, which is proportional to $r^2$.
\end{itemize}
Finally, to manage the memory required by dense 3D prediction, we apply a $4\times$ range, $8\times$ azimuth, and $2\times$ elevation decimation, resulting in (64 range $\times$ 128 azimuth $\times$ 64 elevation) bins, which we output with 1024 $(8 \times 8 \times 8)$ patches. For decimated range-azimuth-elevation grid $r, \theta, \phi$ and 0-1 occupancy $Y^*$, this corresponds to the following loss:
\begin{align}
    \mathcal{L}(&\hat{Y}_{r, \phi, \theta}, Y^*_{r, \phi, \theta}) \nonumber \\
    &= r^2(1.0 + 63.0 Y^*_{r, \phi, \theta})\text{BCE}(\hat{Y}_{r, \phi, \theta}, Y^*_{r, \phi, \theta})
\end{align}

In addition to the test loss, we compute the Chamfer distance (by treating each occupied cell as a point), and the mean absolute error of depth estimates obtained by finding the first occupied cell along the range axis of our range-azimuth-elevation occupancy.

\paragraph{Bird's Eye View (BEV) Occupancy} We use the same binary cross-entropy and Dice loss mixture as \cite{prabhakara2023high}, and output a $256 \times 1024$ range-azimuth polar occupancy grid which corresponds to the native range resolution of the radar and the native azimuth of the Lidar, restricted to forward-facing bins. We output 1024 $(16 \times 16)$ patches.

In addition to the test loss, we also compute the Chamfer distance, using the same procedure as for 3D occupancy.

\paragraph{Semantic Segmentation} We use the $640\times640$ output of \texttt{segformer-b5} \cite{xie2021segformer}, trained on ADE20k \cite{zhou2017scene}, as the ground truth for this task. We aggregate the ADE20k classes into eight broad categories (arranged by index):
\begin{enumerate}[start=0]
    \item \texttt{ceiling}: any structure viewed from below; mostly seen \texttt{indoor}s.
    \item \texttt{flat}: flat, walkable surfaces such as sidewalks and roads. Grass and other vegetation are excluded, and included in \texttt{nature} instead.
    \item \texttt{nature}: vegetation and other natural items such as grass, shrubs, trees, and water.
    \item \texttt{object}: miscellaneous small objects such as furniture which are not included in the \texttt{structure} category.
    \item \texttt{person}: any person who is not inside a vehicle or riding a vehicle.
    \item \texttt{sky}: the sky.
    \item \texttt{structure}: large man-made items such as buildings, fences, and shelters.
    \item \texttt{vehicle}: cars, trucks, buses, and other vehicles.
\end{enumerate}
We output 1024 $(5 \times 5) \times 8$ patches and train using a simple binary cross-entropy loss. Finally, in addition to the test loss, we also calculate the mIOU (mean intersection over union), accuracy, and top-2 accuracy.

\paragraph{Ego-Motion Estimation} We fuse our IMU and Lidar data Cartographer SLAM \cite{hess2016real}, which we project back to the sensor's frame of reference to obtain Ego-Motion ground truth, and manually exclude regions where SLAM failed (typically due to a lack of Lidar features, e.g. bridges or tunnels, totaling $\approx 1\%$ of our dataset).

During training, we first normalize the velocity with respect to the Doppler resolution (i.e. measuring each component in Doppler bins); then, we use the $l_2$ loss
\begin{equation}
    \mathcal{L}_{\text{Ego-Motion}}
    = ||\hat{v} - v^*||_2
    \approx \sqrt{(\hat{v} - v^*)^T(\hat{v} - v^*) + \varepsilon}
\end{equation}
where $\varepsilon = 1.0 \text{ Doppler bins}$ (with a typical magnitude of $16 \leq ||v^*||_2 \leq 32$) is included for numerical stability.

\subsection{Evaluation Procedure}
\label{app:procedure}

\begin{figure}
\centering
\includegraphics[width=0.9\columnwidth]{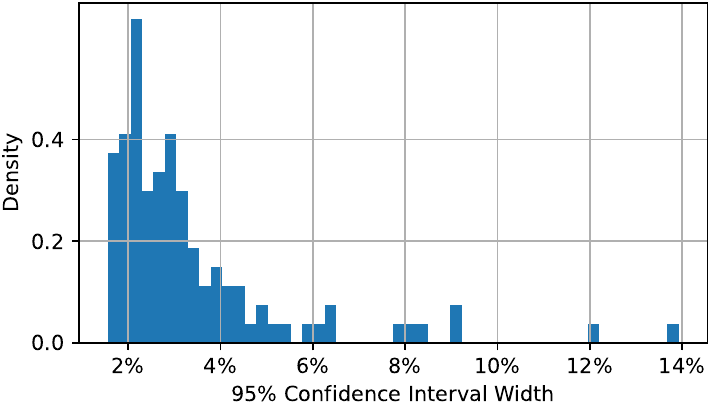}
\vspace{-0.5em}
\caption{Width of 95\% confidence intervals, in percent, relative to the baseline of each ablation, aggregated and plotted as a histogram. \textmd{Using our 4.5 hour (163k frame) test split, we are able to compare methods with a median confidence interval width of 2.6\% (one-sided difference of 1.3\%), with the exact width varying depending on the variance of the underlying comparison.}}
\label{fig:ci_width}
\end{figure}

Following our evaluation procedure, we can measure differences of 1-2\% with high probability (Fig.~\ref{fig:ci_width}); we provide details about this procedure below.

\paragraph{Geo-Split} Within each setting, $\approx$1.5 hours of data are reserved as a test set. In order to control data leakage, we split traces for each setting along natural geographic boundaries:
\begin{itemize}
    \item \texttt{indoor}: since buildings can have duplicated floor plates and other design features between floors or different areas, data was split by building, with the evaluation set consisting of all traces collected from 3 buildings.
    \item \texttt{outdoor}: each trace was collected as a contiguous area on foot; we reserved a set area within a neighborhood that includes various zoning and street types for the test set.
    \item \texttt{bike}: each trace was collected as a round-trip ride from a set origin; two rides from a set range of directions were reserved for the test set.
\end{itemize}

\paragraph{Sample Size Correction} Intuitively, sampling the same signal -- such as radar-lidar-video frames -- with a greater frequency yields diminishing ``information''. Since the standard error of the mean, $\text{SE} = \text{std}(X) / \sqrt{N}$, is defined for $N$ independent and identically distributed samples, we must correct for the \textit{effective sample size} of our test data.

In our analysis, we assuming that changes in model performance imply changes in the underlying data (but not necessarily the converse). This allows us to estimate a lower bound on the effective sample size from each scalar performance metric as \cite{robert1999monte}
\begin{equation}
    N_{\text{eff}} = \frac{N}{1 + 2\sum_{t=1}^\infty \rho_t}
\end{equation}
for autocorrelation $\rho_t$ (where $t$ is the delay). Similar to \cite{huang2024dart}, we approximate the infinite series up to $t = N/2$ and clip negative empirical autocorrelation values $\hat{\rho}_t < 0$ to 0.

\paragraph{Paired z-Test} The actual values of each measured metric have a large inherent variance due to the varying difficulty of the data (e.g., the presence of clutter, dynamics, or other challenging features). As such, the standard error of \textit{absolute} values of each metric is large. Taking advantage of the fact that each model is evaluated on identical test traces with respect to a baseline, and that the performance of each model is highly correlated with its baseline, we instead use a paired z-test, i.e., on the \textit{relative} values of each metric, which mitigates the impact of this variance.

\begin{figure*}
\includegraphics[width=\textwidth]{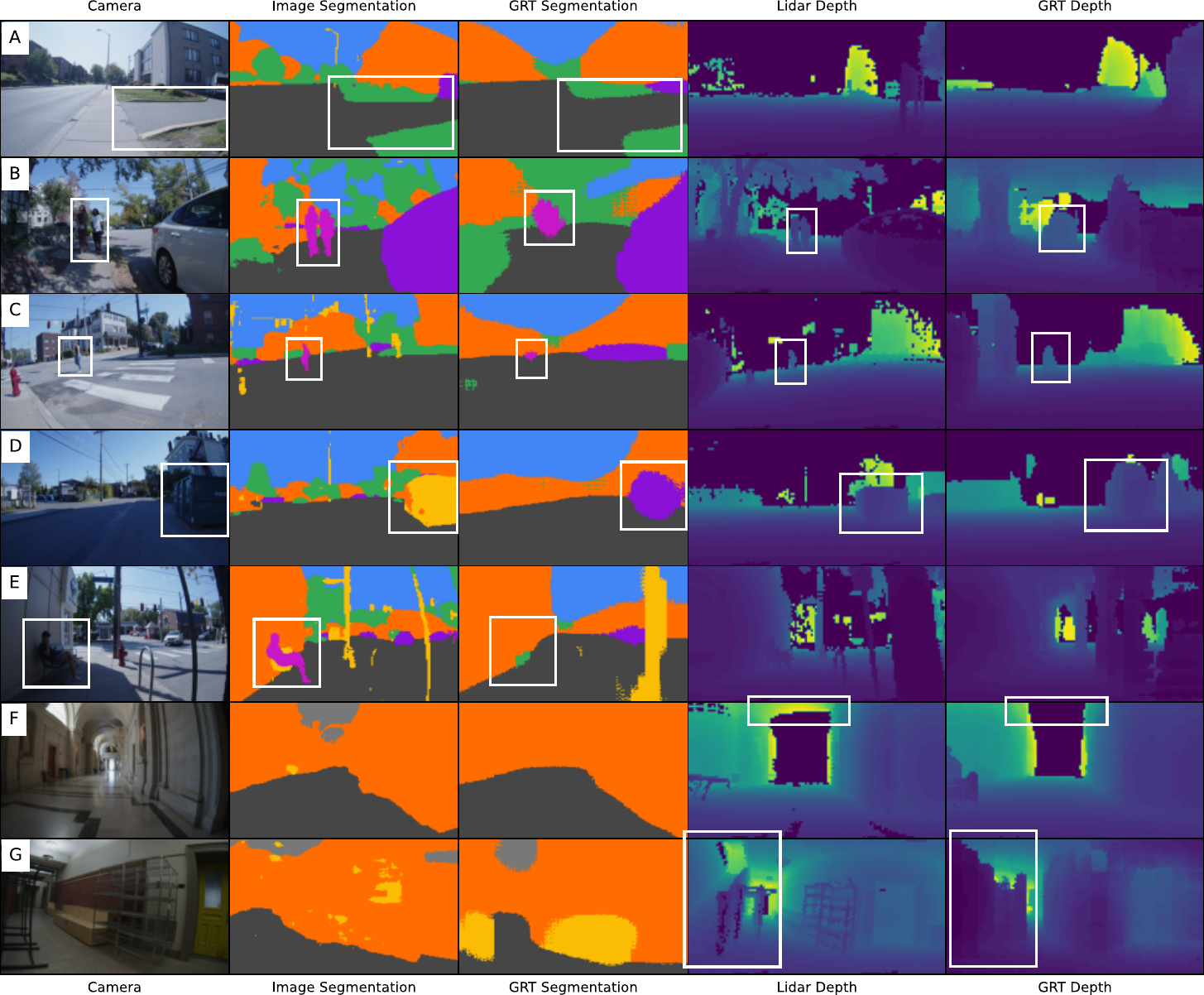}
\vspace{-1.9em}
\caption{\textbf{Select frames from our test set}. Frames are annotated with notable features. Note that the field of view is narrower for the camera ($\approx60^\circ \times 120^\circ$) compared to the Lidar ($90^\circ \times 180^\circ$).}
\label{fig:selected_sample}
\end{figure*}

\subsection{Baseline Details}
\label{app:baseline_details}

Ideally, we would like to compare GRT against off-the-shelf baselines without any modifications. However, due to the lack of standardization in radar hardware and modulations, radar data cubes do not have standard dimensions and aspect ratios; furthermore, since radar data cubes are generally tightly coupled with physical effects arising from hardware and modulation design choices, dimension-normalizing transforms such as cropping and resizing are also not generally valid.

While transformer encoder-decoder architectures can be readily adapted for different input and output dimensions by simply changing the input and output context size (and positional encodings), convolutional and encoder-only architectures must be modified to fit different input and output resolutions. Thus, to run prior baselines on our dataset, we made the a few changes to each baseline.

\paragraph{RadarHD} RadarHD \cite{prabhakara2023high} uses an asymmetric U-net with azimuth-only 2$\times$ upsampling layers to meet the target output resolution, relative to the input. Since RadarHD was originally designed for 512 output azimuth bins instead of the 1024 output azimuth bins in our dataset, we include an additional azimuth upsampling layer, with other layers and dimensions staying the same.

\paragraph{T-FFTRadNet} T-FFTRadNet \cite{giroux2023t} uses a swin transformer encoder with a relatively lightweight convolutional decoder, with some U-net-like skip connections. Since T-FFTRadNet's ``dense'' high-resolution decoder was designed only for cascaded imaging radars, and the decoder for single-chip radar was designed only for sparse outputs, we modified the dense decoder to use the output of the backbone for single-chip radar by increasing the bilinear upsampling size to 4$\times$ in both range and azimuth. Other layers and dimensions are kept the same.

\section{Additional Results}
\label{app:results}

\begin{figure*}
\centering
\includegraphics[width=\textwidth]{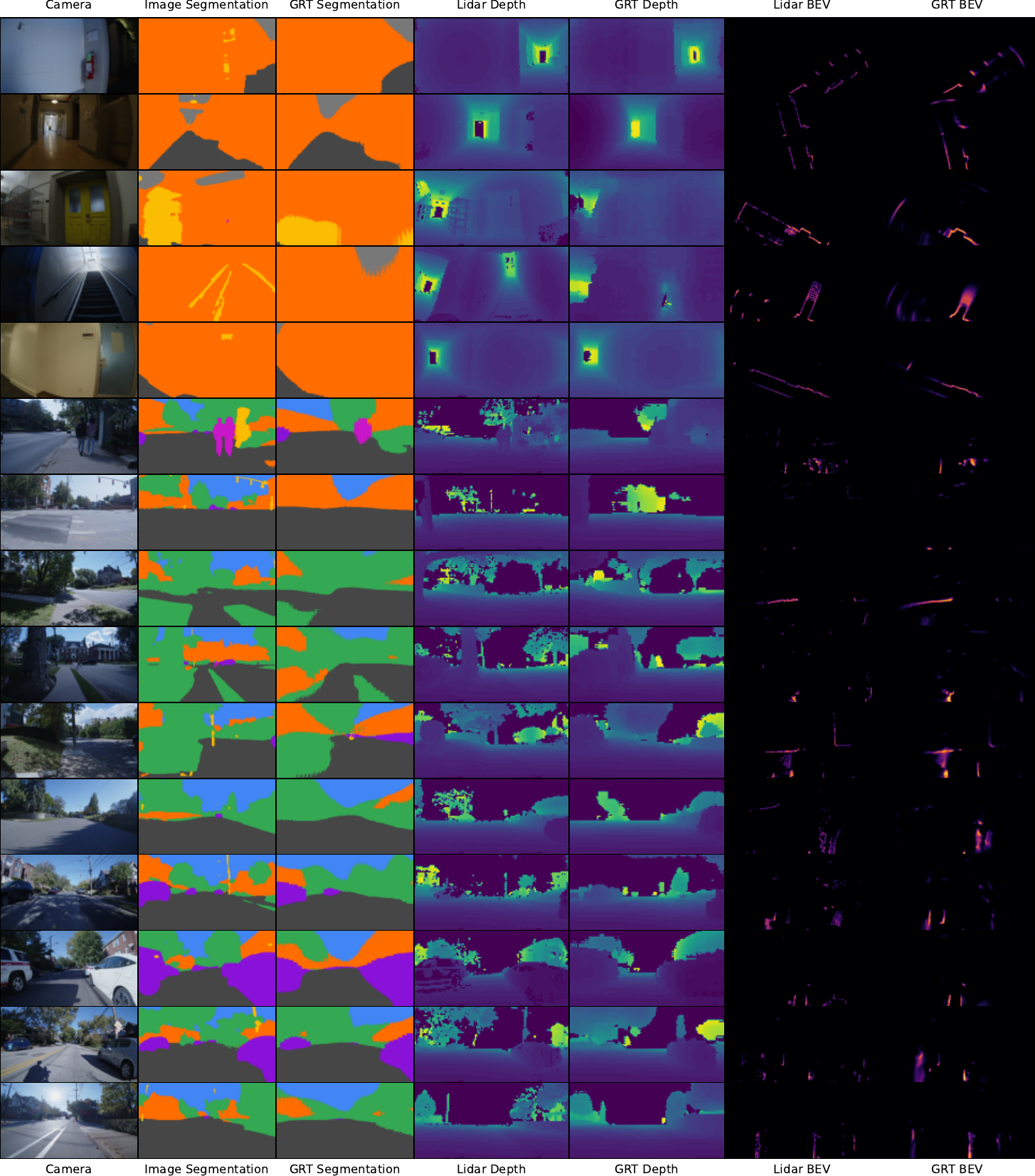}
\caption{\textbf{A random sample of 15 images from the test split of I/Q-1M.} 5 samples are taken from the \texttt{indoor} (top), \texttt{outdoor} (middle), and \texttt{bike} (bottom) settings. All images are generated by \texttt{GRT-small}; the segmentation (left) and BEV (right) outputs are trained and fine-tuned on the full dataset, while the depth output (center) is rendered from the base model's 3D occupancy output.}
\label{fig:random_sample}
\end{figure*}

In this section, we provide sample visualizations (App.~\ref{app:result_samples}) and additional analyses (App.~\ref{app:metrics}-\ref{app:projections}). Note that in addition to the included figures, video examples of our model in action can be found at our project site.

\subsection{Sample Images}
\label{app:result_samples}

To better visualize the capabilities of our model, we provide a range of sample results (Fig.~\ref{fig:selected_sample}), including some cases where our model performs better than expected, failure cases which illustrate the limitations of our approach, and a representative random sample (Fig.~\ref{fig:random_sample}).

\begin{table*}
\caption{Performance metrics (App.~\ref{app:tasks}) of the \texttt{GRT-small} transformer model trained on our base task (3D Occupancy) and fine-tuned on each secondary task with our full dataset (mean with 95\% confidence intervals).}
\footnotesize
\centering
\begin{tabular}{cccccc}
\toprule
Task & Metric & Average & Indoor & Outdoor & Bike \\
\toprule
3D Occupancy & Chamfer & $4.7 \text{ bins} \pm 0.19$ & $0.24\text{ m} \pm 0.02$ & $0.4\text{ m} \pm 0.019$ & $0.38\text{ m} \pm 0.023$ \\
    & Depth & $16 \text{ bins} \pm 0.66$ & $0.62\text{ m} \pm 0.059$ & $1.5\text{ m} \pm 0.09$ & $1.4\text{ m} \pm 0.089$ \\
\hline
Semantic Segmentation & mIOU & $0.69 \pm 0.012$ & $0.78 \pm 0.02$ & $0.63 \pm 0.019$ & $0.69 \pm 0.018$ \\
    & Accuracy & $0.79 \pm 0.0097$ & $0.85 \pm 0.016$ & $0.75 \pm 0.016$ & $0.78 \pm 0.016$ \\
    & Top-2 Accuracy & $0.94 \pm 0.0053$ & $0.97 \pm 0.006$ & $0.92 \pm 0.01$ & $0.93 \pm 0.011$ \\
\hline
BEV Classification & Chamfer & $11\text{ bins} \pm 0.91$ & $0.28\text{ m} \pm 0.027$ & $0.84\text{ m} \pm 0.13$ & $1.3\text{ m} \pm 0.19$ \\
\hline
Ego-Motion Estimation & Speed & $0.95\text{ bins} \pm 0.093$ & $0.025\text{ m/s} \pm 0.0022$ & $0.025\text{ m/s} \pm 0.0024$ & $0.091\text{ m/s} \pm 0.047$ \\
    & Angle & $4.2^\circ \pm 0.46$ & $5.9^\circ \pm 0.54$ & $5^\circ \pm 0.61$ & $1.9^\circ \pm 1.3$ \\
\toprule
\end{tabular}
\label{tab:metrics}
\end{table*}

\begin{figure*}
\centering
\includegraphics[width=\textwidth]{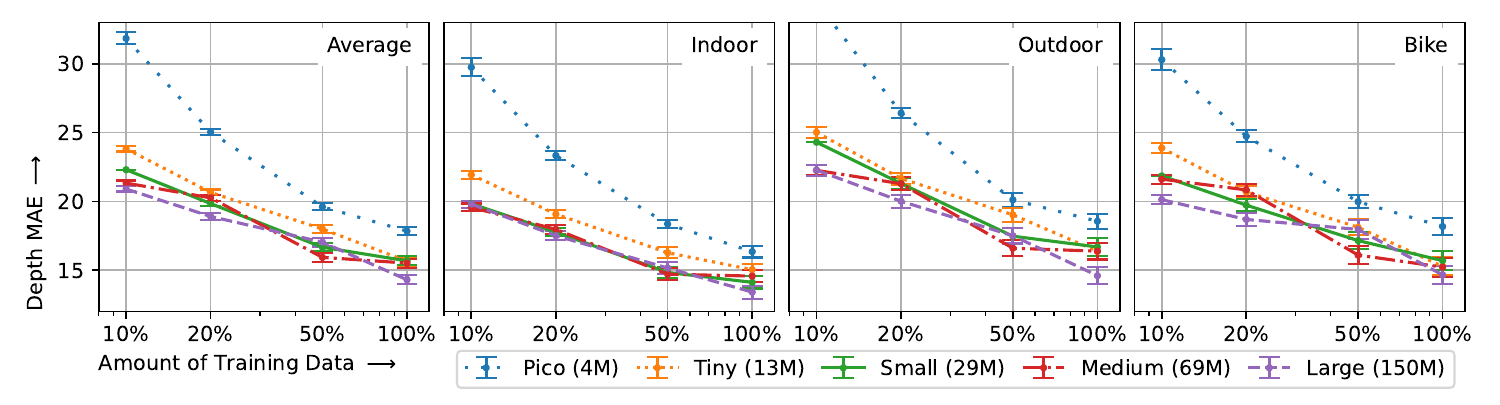}
\vspace{-1.7em}

\includegraphics[width=\textwidth]{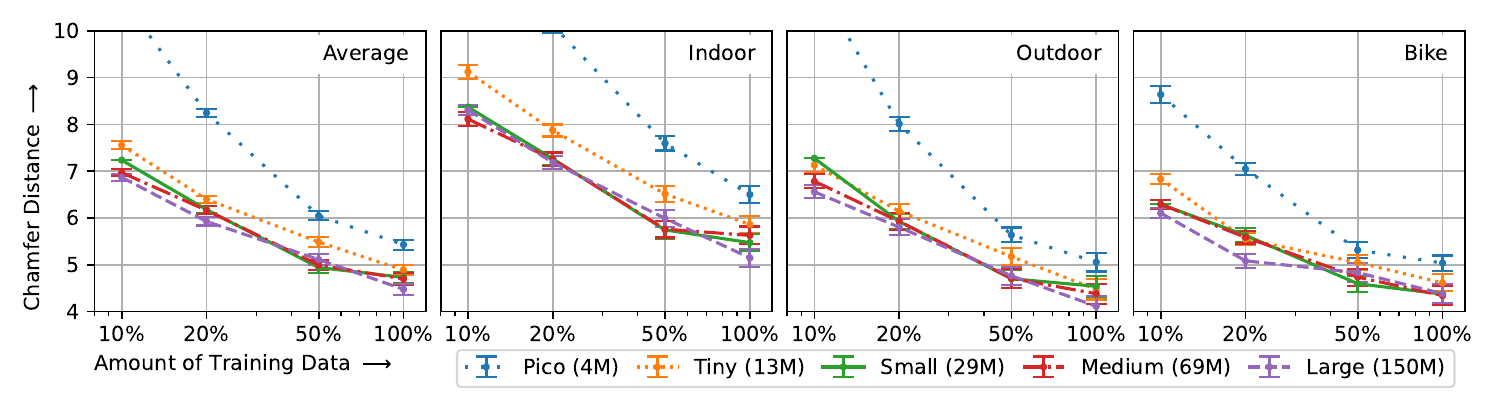}
\vspace{-2em}
\caption{Scaling laws for the \textit{depth} mean absolute error (top) and \textit{chamfer} distance (bottom) metrics, measured in radar range bins (4.4cm \texttt{indoor}, 8.7cm \texttt{outdoor}, \texttt{bike}).}
\label{fig:scaling_depth}
\end{figure*}

\paragraph{Surprising Capabilities} While others have tried mmWave-radar-based semantic segmentation \cite{lai2024enabling}, no prior works attempt to extract high-resolution elevation information for tasks such as semantic segmentation or 3D occupancy classification on such a low-resolution radar. As such, we found it surprising that our model works at all! In our evaluation traces, we also found additional capabilities which further exceeded our expectations:
\begin{itemize}
    \item \textbf{Material Properties}: Despite our radar's low resolution, it is able to correctly label pavement and grass in many cases (Fig.~\ref{fig:selected_sample}, A), likely learning the fact that paved surfaces generally have specular returns, while natural surfaces have diffuse returns.
    \item \textbf{Pedestrians}: the model is able to correctly identify pedestrians in many cases, likely due to the unique micro-doppler signature of people walking (Fig.~\ref{fig:selected_sample}, B-C).
\end{itemize}
Finally, it is worth noting that a radar using GRT's segmentation capability can operate in conditions when cameras cannot such as fog, smoke, and darkness.

\paragraph{Failure Cases} To highlight a few failure cases for GRT:
\begin{itemize}
    \item \textbf{Fine-Grained Classification}: using only a low-resolution radar without any visual or Lidar inputs, a pure radar transformer has no way of differentiating fine-grained classes such as metal dumpsters (in the \texttt{object} class) from the \texttt{vehicle} class (Fig.~\ref{fig:selected_sample}, D).
    \item \textbf{Static People}: without the unique micro-doppler signature associated with walking, our model often fails to detect people who are standing or sitting still (Fig.~\ref{fig:selected_sample}, E).
    \item \textbf{Limited Vertical Resolution}: the vertical field of view of our radar is relatively limited, with a 6dB-beamwidth of $\pm$20\textdegree. Thus, even with Doppler information to help resolve elevation information (beyond the 2 elevation bins measured by our radar), the model cannot reliably estimate regions at the edge of the Lidar or Camera's vertical field of view (Fig.~\ref{fig:selected_sample}, F).
    \item \textbf{Clutter}: when a scene is very cluttered, GRT can fail to resolve individual objects; this generally leads to large hallucinations (Fig.~\ref{fig:selected_sample}, B, G).
\end{itemize}

\begin{figure*}
\centering
\includegraphics[width=\textwidth]{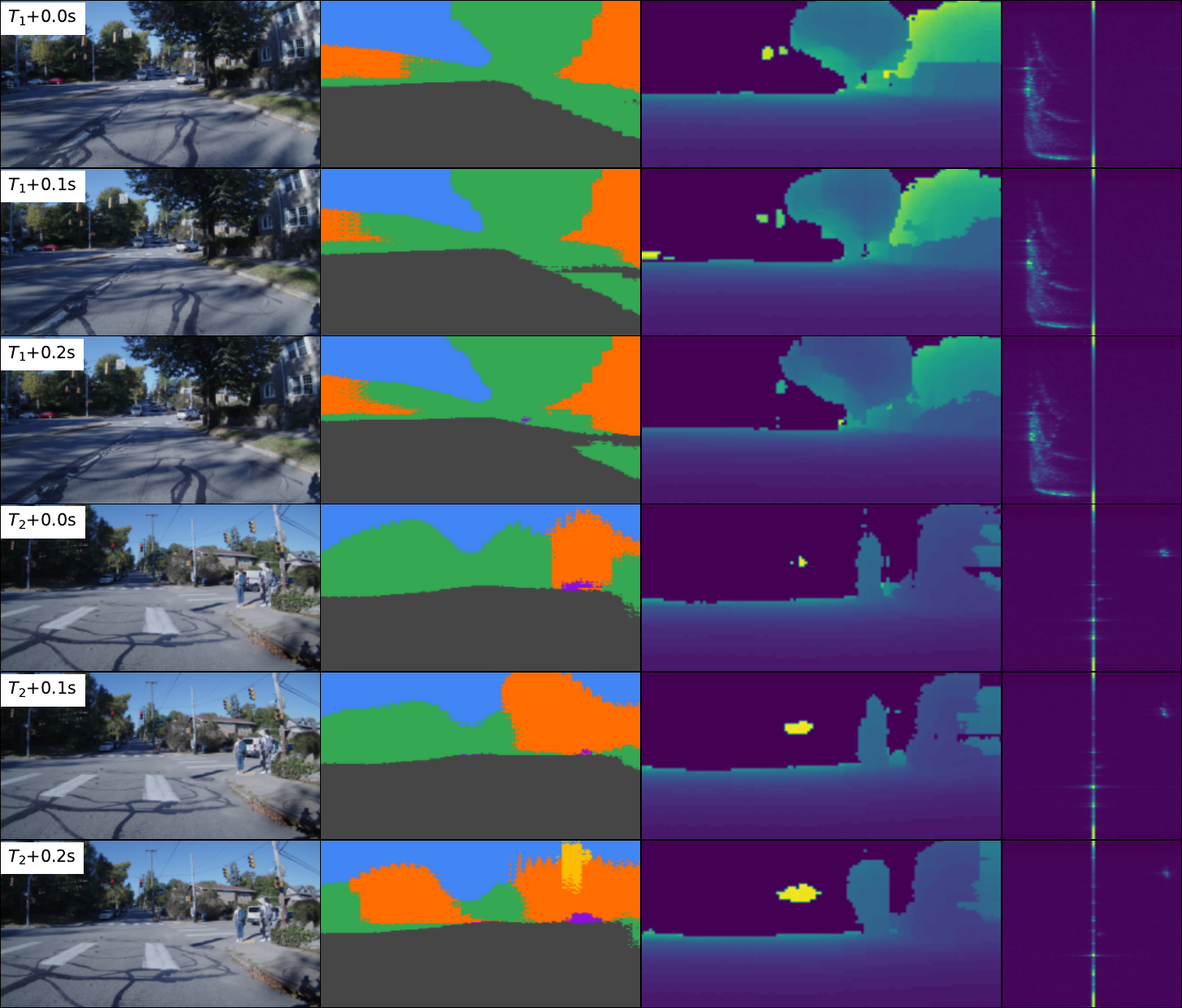}
\vspace{-1.7em}
\caption{\textbf{Sample sequences of three consecutive frames before (top) and after (bottom) stopping at a red light.} Video frames are provided for reference (left), with GRT's semantic segmentation (center left) and depth (center right) outputs across the two sequences along with the range-Doppler spectrum. After stopping, the Doppler spectrum (horizontal axis; right) collapses to a single Doppler bin, resulting in significantly decreased information available to the model. This manifests as noisier (as seen by larger frame-to-frame variations and hallucinations) and less accurate predictions by the model.}
\label{fig:doppler-example}
\end{figure*}

\subsection{Absolute Metrics}
\label{app:metrics}

Since each task has a number of possible metrics (which are not always aligned), we generally report metrics as relative test losses to best capture the performance \textit{of the model} in its ability to fit the target loss.

\paragraph{Key Metrics} As an absolute reference, we calculated a range of common performance metrics for each objective (Table~\ref{tab:metrics}) as described previously (App.~\ref{app:tasks}). For these metrics, SI units are reported where applicable; distance and speed metrics are normalized by range and Doppler resolutions, respectively, when aggregating over settings with different radar configurations.

\paragraph{Absolute Scaling Laws} As an alternate version to Fig.~\ref{fig:scaling}, we also measured scaling laws with respect to the \textit{depth} and \textit{chamfer} metrics (Fig.~\ref{fig:scaling_depth}); while somewhat noisier, the same general trend can be seen. Note that this noise is also why we compare loss metrics in our scaling laws and ablations since it serves as a more direct measure of relative ``learning'' performance.

\subsection{Impact of Doppler}
\label{app:doppler}

\begin{figure}
\centering
\includegraphics[width=0.9\columnwidth]{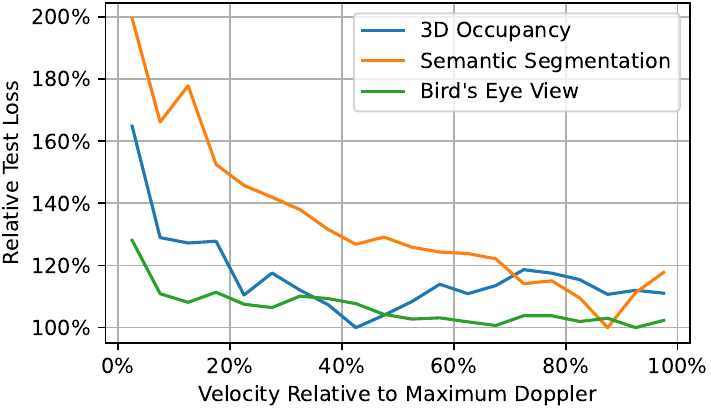}
\vspace{-0.5em}
\caption{Relative test losses binned by the speed of the data collection rig (20 equal 5\% bins) on the \texttt{bike} subset; \textmd{due to less available Doppler information, lower speeds are associated with higher losses for each of our tasks.}}
\label{fig:speed_loss}
\end{figure}

To further illustrate the impact of Doppler, we measured the test loss of our objectives (other than Ego-Motion estimation), binned against the sensor speed (Fig.~\ref{fig:speed_loss}). When the sensor's speed is low relative to its maximum Doppler, it captures less Doppler information due to our radar's fixed Doppler resolution, leading to degraded performance. This can also be seen qualitatively: when the data capture rig stops, the GRT model's predictions become noisier, less sharp, and tend to show blocky artifacts aligned with the output patch size (Fig.~\ref{fig:doppler-example}).

\subsection{Scaling Law Projections}
\label{app:projections}

To motivate future work scaling data collection and training for single-chip radar models using 4D data cubes, we run a suite of scaling law experiments to obtain rough, order-of-magnitude estimates for the data requirements of ``fully'' training GRT foundational model (Sec.~\ref{sec:how_much_data}) of a similar size to the relatively small (by vision standards) models which we trained. In this section, we describe additional details and assumptions behind our two methods of estimation.

\paragraph{Extending the Scaling Law} In order to estimate data requirements from our scaling law, we start from the assumption that data scaling will always be at most logarithmic. This is based on the intuition that increasing the size of the dataset will always have diminishing returns (in turn at a diminishing rate), which is consistently observed in other scaling experiments \cite{zhai2022scaling}.

We then train a network only on the test set, without any augmentations; we reason that this provides a lower bound on the achievable test loss (due to random, unpredictable noise in the dataset) for a given architecture, given that the model is not large enough to memorize the test set pixel-for-pixel. Note that this also provides an improvement over a naive lower bound from the fact that $\mathcal{L} > 0$.

Combining these two implies that data scaling can be logarithmic for increasing dataset size up to at most 100$\times$ our current dataset size, which we believe represents a reasonable order-of-magnitude estimate for the data requirements for a ``fully trained'' radar foundational model.

\paragraph{Training Curve Patterns}

\begin{figure}
\centering
\includegraphics[width=\columnwidth]{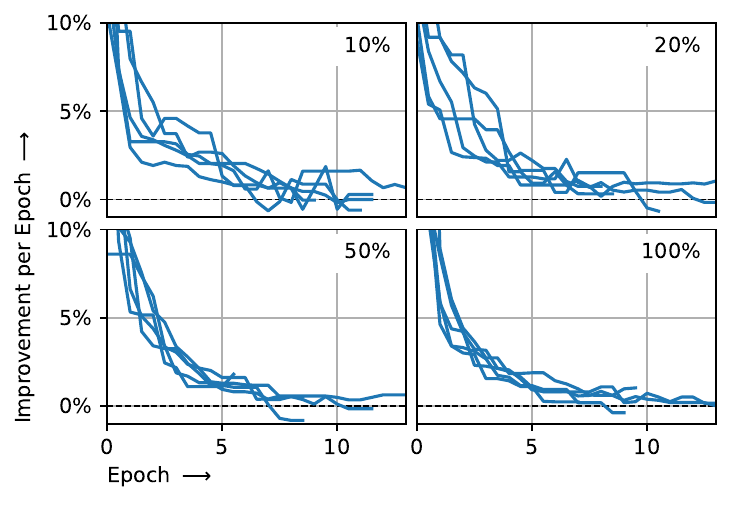}
\vspace{-2.5em}
\caption{Validation loss improvement per epoch, measured every checkpoint (2 checkpoints/epoch), and smoothed with a 5-checkpoint median filter. Each line refers to a different model (with different sizes); models are separated by dataset size. Our models consistently tend to stop improving (in validation loss) after around 10 epochs of training.}
\label{fig:epoch_trend}
\end{figure}

In our experiments, we observe that all models tend to stop improving with respect to validation loss after approximately 10 epochs (Fig.~\ref{fig:epoch_trend}). This gives a further avenue for projections: assuming that the informational ``value'' of a radar frame is roughly equivalent to an image, we can take rough numbers for the typical number of samples seen used to train a vision transformer and translate this to data requirements for a radar foundational model.

Note that this assumption of informational equivalence is also quite rough. Unlike vision transformers, which are typically trained on \textit{independent} images scraped from the internet, GRT is trained using \textit{dependent} frames sampled from a time-series of sensor data, decreasing the relative information density of radar time-series data. On the other hand, while vision transformers typically use sparse feedback signals such as image-caption \cite{radford2021learning} or image-label \cite{zhai2022scaling} pairs, GRT is trained using dense feedback in the form of a 3D occupancy grid (App.~\ref{app:tasks}). In principle, this increases the relative information density of radar-Lidar training pairs. Our projection therefore assumes that these factors roughly balance out within an order of magnitude.

\end{document}